%% file: terrabench_arxiv.tex
\documentclass{article}
\usepackage[utf8]{inputenc}
\usepackage[table]{xcolor}
\usepackage{booktabs}
\usepackage{multirow}
\usepackage{graphicx}
\definecolor{sectiongray}{RGB}{240,242,247}
\usepackage{booktabs}
\usepackage{xcolor}
\usepackage{pifont}
\usepackage{threeparttable}
\usepackage{graphicx}
\usepackage{caption}
\usepackage{svg}
\usepackage{subcaption}
\usepackage{float}

\newcommand{\cmark}{\textcolor{green!60!black}{\ding{51}}}
\newcommand{\xmark}{\textcolor{red!75!black}{\ding{55}}}
\newcommand{\pmark}{\textcolor{orange!85!black}{$\triangle$}}

\usepackage[eandd, preprint, nonanonymous]{neurips_2026}

\usepackage[preprint]{neurips_2026}

\usepackage[utf8]{inputenc} 
\usepackage[T1]{fontenc}    
\usepackage{hyperref}       
\usepackage{url}            
\usepackage{booktabs}       
\usepackage{amsmath}        
\usepackage{amsfonts}       
\usepackage{nicefrac}       
\usepackage{microtype}      
\usepackage{xcolor}         

\title{TerraBench: Can Agents Reason Over \\Heterogeneous Earth-System Data?}

\author{%
  \textbf{Dat Tien Nguyen}$^{1}$ \quad
  \textbf{Thao Nguyen}$^{1}$ \quad
  \textbf{Fadillah Adamsyah Maani}$^{1}$ \\
  \textbf{Huy M. Le}$^{1}$ \quad
  \textbf{Muhammad Umer Sheikh}$^{1}$ \quad
  \textbf{Numan Saeed}$^{1}$ \\
  \textbf{Muhammad Haris Khan}$^{1}$ \quad
  \textbf{Salman Khan}$^{1,2}$ \\
  $^{1}$Mohamed bin Zayed University of Artificial Intelligence, Abu Dhabi, UAE\\
  $^{2}$ADIA Lab, Abu Dhabi, UAE
}

\begin{document}

\maketitle
\begin{abstract}

Climate and environmental decision-making increasingly requires reasoning across heterogeneous inputs, including gridded physical data, satellite imagery, geospatial context, and simulator outputs. Weather and climate foundation models can forecast well, but do not reason interactively in language, while large language models (LLMs) reason in language but cannot operate directly on high-dimensional Earth-system data. As a result, real scientific workflows in Earth-science remain underserved. We introduce TerraBench, a benchmark for grounded Earth-science reasoning, built on TerraAgent, a ReAct-style executable framework that interleaves reasoning, tool calls, and observations to couple LLM planning with scientific tools for environmental retrieval, geospatial processing, simulation, and artifact-backed computation. TerraBench unifies analysis of Earth observation imagery, gridded data, GIS reasoning and simulation in a single executable interface, whereas prior benchmarks isolate these capabilities into narrow individual tasks. It is also the first in this space to pair process-level tool-use metrics with tolerance-aware numeric scoring. The benchmark comprises 403 extensive agentic tasks across three tracks (Fundamentals, Simulator-Grounded, and Document-Grounded Verification) and eight application domains with $\sim$24,500 verified execution steps. The benchmark is rigorously filtered and refined: only 50.9\% of proposed samples were retained after human review, and 74.4\% of those required multiple execution passes to finalize. Empirically, TerraBench exposes the limitations of today’s strongest models. The top-performing frontier model (Claude Sonnet 4.6) reaches only 59.2 ToolUseScore and 22.9 Hit@tol, while the strongest open-weight model (Qwen3.5-35B) trails at 40.0 and 5.9. This gap is driven mainly by argument and numeric grounding failures rather than tool selection errors, with over 84\% of numerical answers falling outside acceptable error margins across all models. These results indicate that reliable Earth-science agents must go beyond tool access to coordinate heterogeneous workflows, parameterize tools precisely, and preserve artifact provenance. Our code is available on \href{https://github.com/Takerdat23/TerraBench}{TerraBench}.

\end{abstract}

\begin{figure}[htbp]
    \centering
    \includegraphics[width=0.85\textwidth]{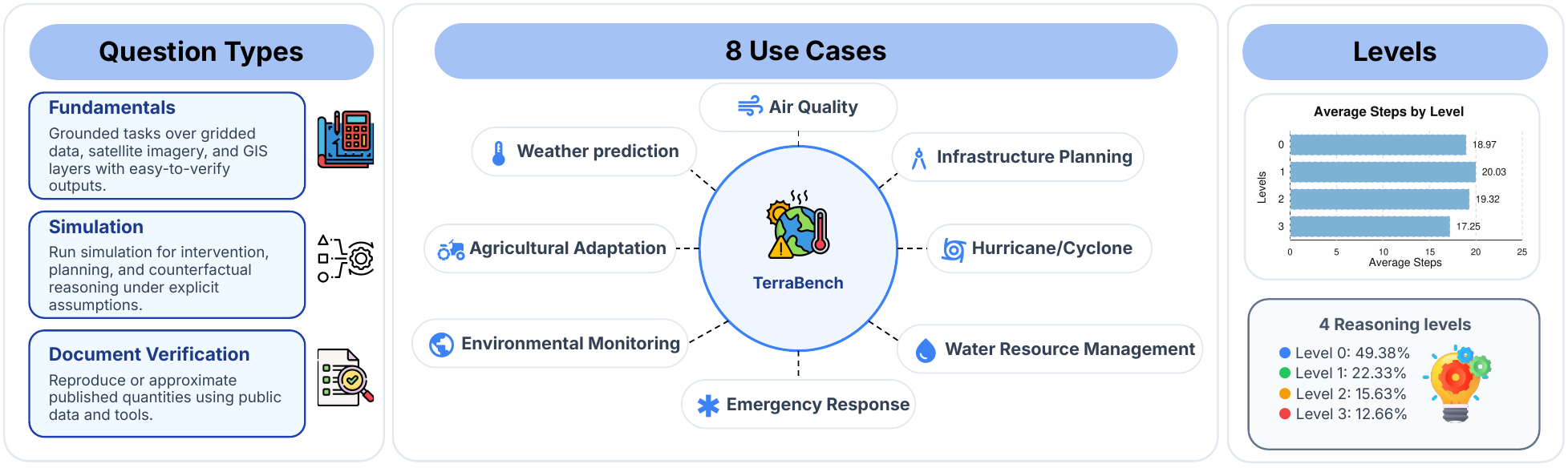}
    \caption{\textbf{TerraBench Overview.}
TerraBench is organized around three task tracks, eight application domains, and four reasoning levels. (Left) \textit{Fundamentals} for directly verifiable multimodal execution tasks, \textit{Simulator-Grounded} for intervention and counterfactual simulation workflows, and \textit{Document-Grounded Verification} for reconstructing or approximating published scientific quantities. (Center) The eight use-cases include weather, air quality, emergency response, water resources, agriculture, infrastructure, environmental monitoring, and hurricane/cyclone analysis. (Right) Reasoning-level distribution and the average number of execution steps per level, showing the benchmark’s depth.}
    \label{fig:overview}
\end{figure}

\section{Introduction}

Large language models (LLMs) have motivated the development of autonomous agents for multi-step scientific workflows \cite{react,gaia,agentx}. In the environmental sciences, systems such as Zephyrus show that LLMs can interact with meteorological datasets and simulators \cite{varambally2025zephyrus}. However, current weather agents and Earth observation (EO) benchmarks like OpenEarthAgent \cite{openearthagent} and ThinkGeo \cite{thinkgeo} still rarely evaluate whether an agent can jointly reason over satellite imagery, gridded environmental data, GIS context, simulation outputs, and document-grounded evidence under a unified, auditable workflow. This gap is critical: without unified evaluation, claims about Earth-science agent capabilities cannot be compared across systems, and numeric-tolerance failures go undetected because existing benchmarks \cite{openearthagent, thinkgeo,sheikh2026gca, varambally2025zephyrus} primarily measure tool-trace alignment. Practical climate and environmental questions inherently require the tool-grounded fusion of these heterogeneous sources, such as ERA5/CMIP6/C3S physical datasets and OpenStreetMap layers. 

To address this gap, we introduce TerraBench, a benchmark for grounded Earth-science reasoning, together with TerraAgent, an executable tool-augmented framework. TerraAgent is a ReAct-style tool-augmented framework that plans over climate questions, invokes domain-specialized scientific tools, materializes intermediate artifacts, and returns structured, provenance-bearing outputs. Built on this framework, TerraBench contains 403 executable complex tasks across 3 tracks, 8 application domains, 77 sub-tools and with nearly 24,500 execution steps. Unlike static QA datasets, each item is a high-information executable benchmark program with a structured output contract, canonical reasoning trace, tool observations, supporting artifacts, and verified final answer. TerraBench explicitly separates process-level and outcome-level evaluation. Empirically, the benchmark remains challenging for both frontier and open models: Claude Sonnet 4.6 achieves 59.22 ToolUseScore, 28.44 NumScore, and 22.88 Hit@tol, while the strongest reported open-weight model, Qwen3.5-35B, reaches 39.95, 7.49, and 5.89, respectively. Track-level and failure analyses further show that simulator-grounded tasks are especially difficult and that dominant failure modes include wrong argument values, wrong tool ordering, and numeric misses beyond tolerance.

In summary, our contributions are:
\begin{itemize}
    \item \textbf{A new benchmark:} TerraBench, a benchmark of 403 long-horizon executable tasks with nearly 24,500 curated steps across three tracks and eight application domains, the first to unify Earth observation imagery, gridded environmental data, GIS reasoning, deterministic simulation, and document-grounded verification under a single executable interface.
    \item \textbf{An executable framework:} TerraAgent, a ReAct-style executable framework with 77 scientific sub-tools that produces auditable reasoning traces and artifact-backed outputs (NetCDF, GeoTIFF, CSV, PNG), enabling reproducible workflows across baseline, intervention, and counterfactual scenarios.
    \item \textbf{A fine-grained evaluation protocol:} A tolerance-aware evaluation protocol that decouples process-level tool-use proficiency from final-answer numerical correctness. Our results show a systematic gap (e.g., 59.2 ToolUseScore vs. 22.9 Hit@tol for the strongest frontier model) that prior tool-trace-only evaluation cannot detect.
\end{itemize}

\begin{table*}[t]
\centering
\footnotesize
\captionsetup{width=\textwidth}
\renewcommand{\arraystretch}{1.12}
\begin{threeparttable}
\label{tab:benchmark_comparison_revised}

\begin{tabular}{lccccccccccc}
\toprule
\textbf{Benchmark} 
& \textbf{\shortstack{Deployed\\tools}} 
& \textbf{\shortstack{EO\\imagery}} 
& \textbf{\shortstack{Env.\\data}} 
& \textbf{\shortstack{GIS\\ops}} 
& \textbf{\shortstack{Simulator\\tasks}} 
& \textbf{\shortstack{Numeric\\eval.}} 
& \textbf{\shortstack{Real-\\world}} 
& \textbf{\shortstack{Hybrid\\annot.}} 
\\
\midrule
ThinkGeo~\cite{thinkgeo}             & \cmark & \cmark & \xmark & \pmark & \xmark & \xmark & \cmark & \cmark \\
GeoBenchX~\cite{geobenchx}           & \cmark & \xmark & \xmark & \cmark & \xmark & \xmark & \pmark & \xmark \\
OpenEarthAgent~\cite{openearthagent} & \cmark & \cmark & \xmark & \cmark & \xmark & \xmark & \cmark & \pmark  \\
GeoMMBench~\cite{geommbench}         & \pmark & \cmark & \xmark & \cmark & \xmark & \xmark & \xmark & \xmark \\
ZephyrusBench~\cite{varambally2025zephyrus}   & \cmark & \xmark & \cmark & \pmark & \cmark & \pmark & \pmark & \pmark \\
GCA~\cite{sheikh2026gca}              & \cmark & \cmark & \cmark & \cmark & \pmark & \cmark & \cmark & \pmark \\
GAIA~\cite{gaia}                     & \xmark & \xmark & \xmark & \xmark & \xmark & \xmark & \cmark & \xmark \\
Agent-X~\cite{agentx}                & \cmark & \xmark & \xmark & \xmark & \xmark & \xmark & \cmark & \cmark \\
GTA~\cite{gta}                       & \cmark & \xmark & \xmark & \xmark & \xmark & \xmark & \cmark & \xmark \\
\midrule
\textbf{Ours}                        & \textbf{\cmark} & \textbf{\cmark} & \textbf{\cmark} & \textbf{\cmark} & \textbf{\cmark} & \textbf{\cmark} & \textbf{\cmark} & \textbf{\cmark}  \\
\bottomrule
\end{tabular}

\caption{\textbf{Comparison with related geospatial, EO, climate, and general agentic benchmarks.}
Columns indicate whether each benchmark includes a deployed executable tool environment, Earth observation (EO) imagery, gridded environmental (Env.) data, geospatial operations (GIS ops), simulator/counterfactual tasks, structured numeric evaluation, real-world tasks/queries, and hybrid human+AI/semi-automated annotation (Hybrid annot.). \cmark =yes, \pmark =partial, \xmark =no.}
\end{threeparttable}
\end{table*}

\section{Related Work}

\textbf{Multimodal Agentic Benchmarks.} 
Recent LLM evaluation has increasingly shifted from static question answering toward agentic tasks that require tool use, long-horizon execution, and evidence-grounded decision-making. ReAct introduced an interleaved reasoning--acting--observing paradigm \cite{react}, and later systems such as HuggingGPT, Visual ChatGPT, and MM-ReAct extended tool-augmented reasoning to broader and multimodal tool ecosystems \cite{hugginggpt,visualchatgpt,mmreact}. Benchmark efforts have followed a similar trajectory: ToolBench, API-Bank, and StableToolBench focus on API/tool invocation \cite{toolbench,apibank,stabletoolbench}; VisualWebArena and MLGym emphasize realistic long-horizon execution \cite{visualwebarena,mlgym}; and general agentic benchmarks such as GAIA, GTA, and Agent-X evaluate real-world assistant tasks, deployed tool use, and multimodal step-level reasoning \cite{gaia,gta,agentx}. These benchmarks are closely aligned with our evaluation philosophy, especially in their focus on tool grounding and process diagnostics. However, they do not target the climate-specific combination of EO imagery, gridded environmental data, GIS reasoning, simulation, and document-grounded verification. TerraBench fills this gap by adapting agentic benchmark design to grounded Earth-science workflows under a unified three-track task interface.

\textbf{Earth-Science Agentic Benchmarks.}
Recent work has introduced multimodal datasets and agentic systems for weather, climate, EO, and geospatial reasoning. Terra, ClimateIQA, WeatherQA, and CLLMate combine environmental observations with text, imagery, remote sensing, or ERA5-based event information \cite{chen2024terra,10.1145/3711896.3737406_climateIQA,ma2024weatherqa,li-etal-2025-cllmate}, while UnivEARTH and GeoHOP move toward document-verified EO questions and hierarchical geospatial reasoning \cite{univearth,geohop}. In parallel, AutoClimDS, Zephyrus, and HVR-Met connect LLMs with climate datasets, forecasting tools, simulators, and meteorological diagnosis workflows \cite{autoclimds2025,varambally2025zephyrus,nguyen2024scaling,davenport2026jcm,tang2026hvrmet}. ThinkGeo, GeoBenchX, OpenEarthAgent, and Earth-Agent demonstrate tool-grounded EO and GIS reasoning \cite{thinkgeo,geobenchx,openearthagent,earthagent}. However, existing work typically specializes in EO perception, weather reasoning, geospatial tool use, or climate data access. TerraBench instead unifies EO imagery, gridded environmental data, GIS reasoning, simulation, and document-grounded verification under one benchmark. It also extends prior evaluation practice by pairing process-level tool-use metrics with tolerance-aware numeric answer scoring, rather than relying only on tool-trace evaluation or LLM-as-judge final-answer assessment. This is particularly important for Earth-science workflows, where final outputs are heterogeneous numerical quantities across execution-grounded, simulator-grounded, and document-grounded tasks.

\begin{figure}[t]
    \centering
    \includegraphics[width=0.8\linewidth]{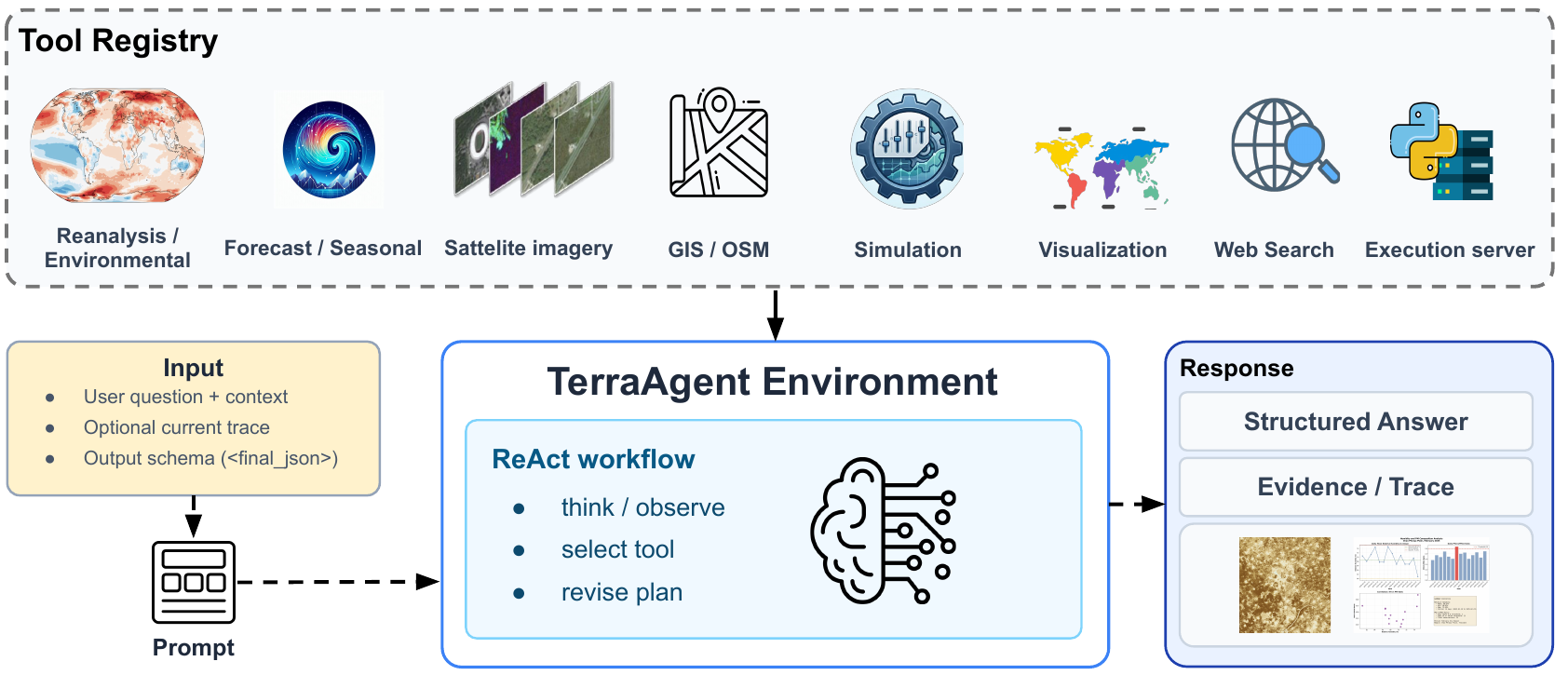}
    \caption{\textbf{Agentic framework overview.} TerraAgent takes a user question, optional prior trace, and structured output schema, then executes a tool-grounded workflow over a domain-organized scientific tool registry. The framework returns both a structured final answer and a provenance-bearing evidence trace with supporting artifacts.}
    \label{fig:framework_overview}
\end{figure}

\section{TerraAgent Framework}
\label{sec:framework}

\textbf{TerraAgent} is the executable framework underlying TerraBench. It turns open-ended climate and environmental questions into auditable, tool-grounded scientific workflows. Given a user question, context, and a structured output requirement, TerraAgent plans a workflow, invokes domain-specific tools, records intermediate observations, materializes artifacts, and returns both a structured answer and an evidence trace. The central design principle is to separate language-based planning from scientific execution: the model coordinates the workflow, while quantitative outputs must come from data retrieval, geospatial processing, forecasting, simulation, visualization, or explicit computation rather than unsupported language-model generation. As shown in Figure~\ref{fig:framework_overview}, TerraAgent is built around a domain-organized tool registry covering the capabilities required by TerraBench: reanalysis and environmental-data retrieval, forecast and seasonal products, satellite/EO processing, GIS and OpenStreetMap analysis, deterministic simulation, visualization, web search, and an execution server for auxiliary scientific computation. The forecasting stack includes learned forecast wrappers such as Pangu-Weather~\cite{pangu} and Aurora~\cite{aurora/nature/BodnarBLSABGRWD25}, together with seasonal and verification utilities. The simulation stack contains deterministic simulators spanning diverse impact domains. These include AquaCrop crop-water response, DSSAT crop systems, CLIMADA event impact estimation, UTCI heat-stress assessment, EnergyPlus building demand analysis, and SUMO traffic disruption analysis~\cite{aquacrop, dssat, climada, utci, energyplus, sumo}. 
Overall, these tools support comprehensive workflows across all Earth-System data.

\section{Benchmark Creation}

\begin{figure}[H]
    \centering
    \includegraphics[width=0.7\linewidth]{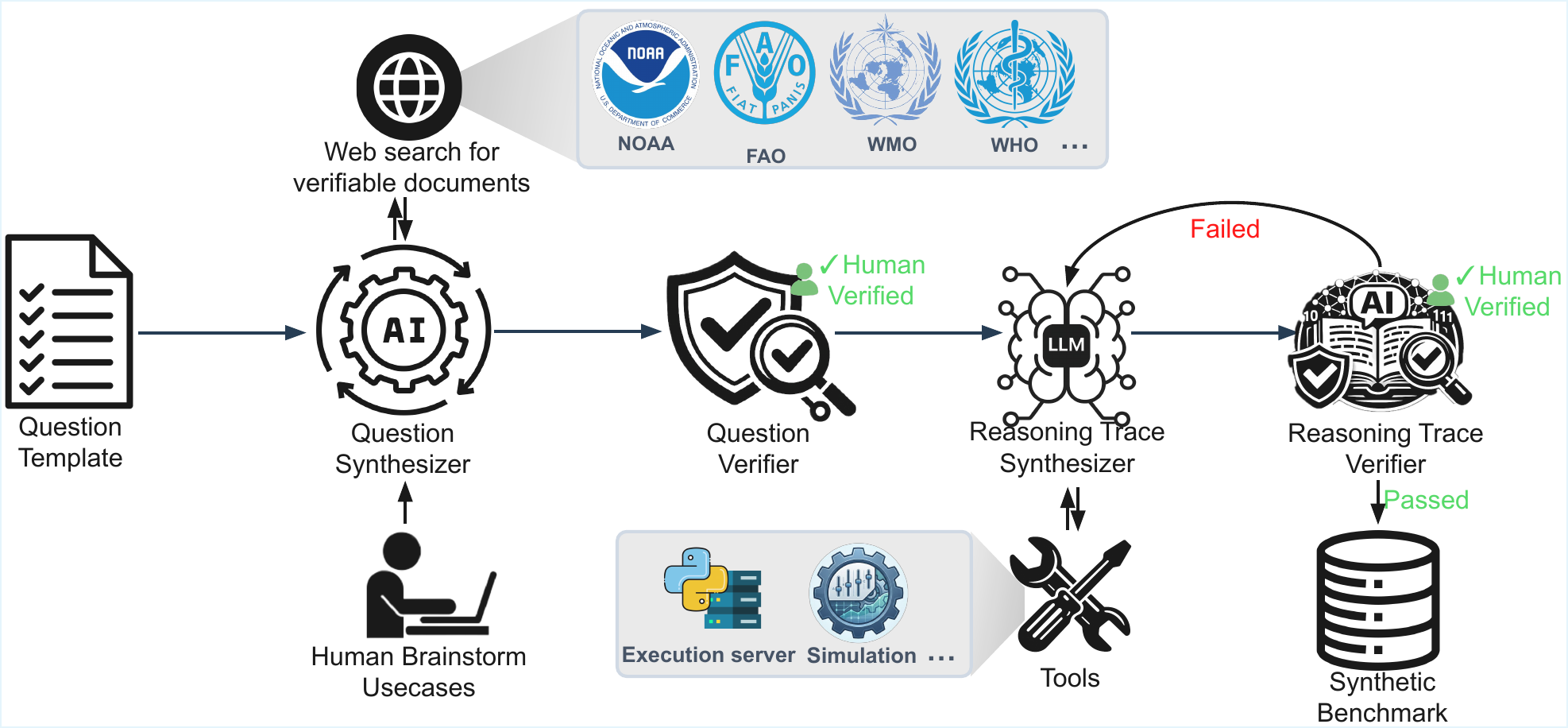}
    \caption{Annotation pipeline for TerraBench. All samples are manually verified for accuracy.}
    \label{fig:data_curate}
\end{figure}

\subsection{Task Taxonomy, Schema, and Causal Reasoning Levels}
\label{sec:task_taxonomy}

TerraBench spans three tracks: Fundamentals (evaluating deterministic execution), Simulator-Grounded (deterministic simulation), and Document-Grounded Verification (scientific document anchors). Covering eight environmental domains, the benchmark replaces static question-answer pairs with executable benchmark programs. Each item shares a common interface—a natural-language question paired with a structured output contract—and includes a canonical reasoning trace, supporting artifacts, and a verified final answer. We additionally annotate items across four reasoning levels. Levels 1--3 adapt Pearl’s causal hierarchy~\cite{pearl2009causality}: associational (Level 1), interventional (Level 2), and counterfactual/retrospective reasoning (Level 3). We introduce Level 0 as a benchmark-specific observational grounding layer. While not strictly causal, Level 0 captures the prerequisite capabilities required before meaningful causal analysis can occur, forcing agents to resolve spatiotemporal coordinates, align heterogeneous data grids, and compute observed quantities. Crucially, these levels denote causal depth rather than a strict monotonic difficulty ordering. Although conceptually simpler, Level 0 remains operationally challenging for current agents; our results demonstrate that even strong models frequently fail on observational grounding due to incorrect tool parameterization, workflow sequence errors, or numeric misses beyond tolerance.

\subsection{Benchmark Annotation and Review Protocol}
\label{sec:benchmark_annotation}

\begin{figure}[t]
    \centering
    \begin{minipage}[t]{0.5\textwidth}
        \centering
        \includegraphics[width=\linewidth]{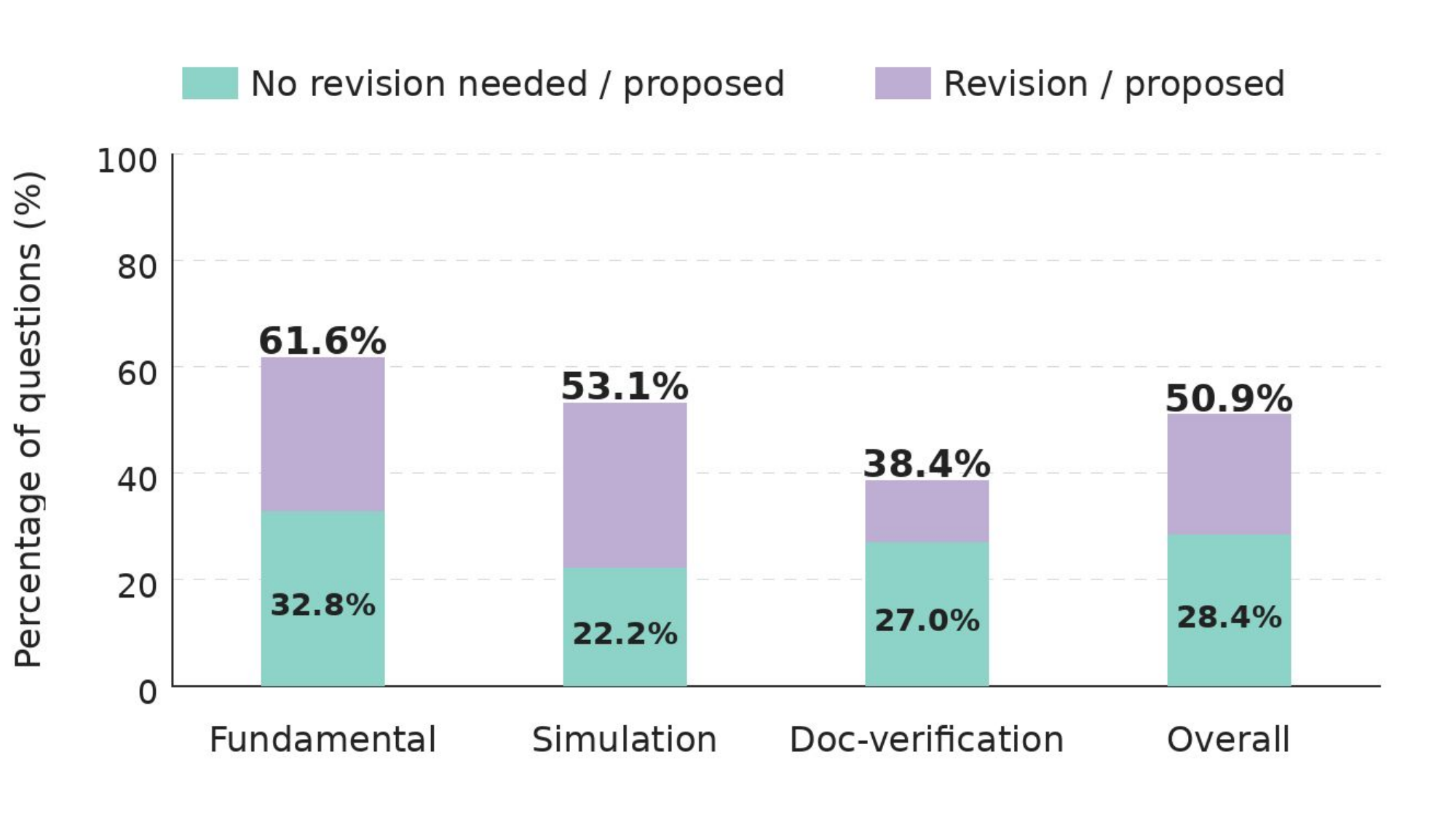}
        {\small (a)}
    \end{minipage}
    \begin{minipage}[t]{0.48\textwidth}
        \centering
        \includegraphics[width=\linewidth]{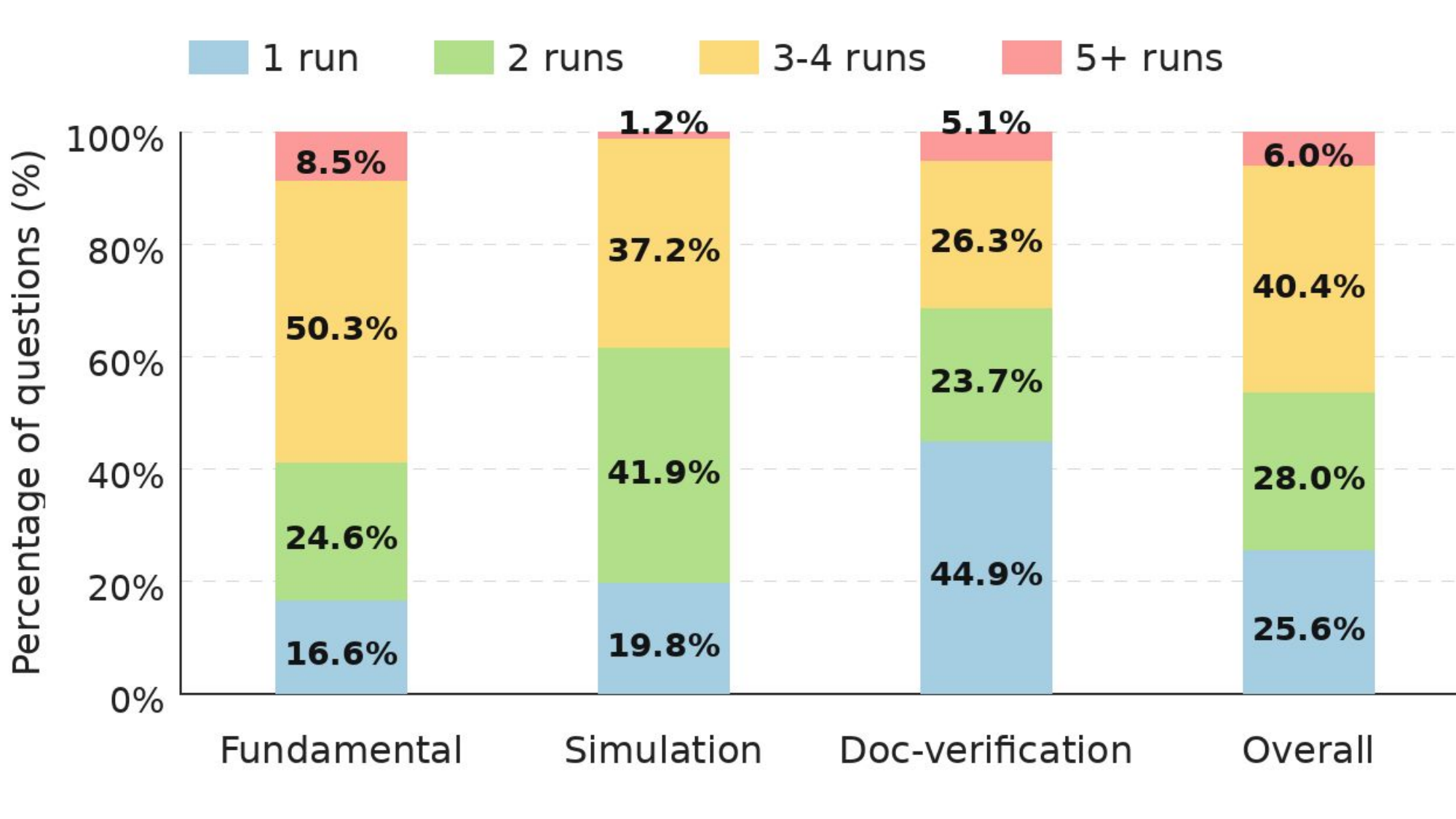}
        {\small (b)}
    \end{minipage}
    \caption{\textbf{Annotation selectivity and iteration burden.}
(a) Share of proposed questions retained in each track, split into items accepted without revision and items accepted after revision; the unfilled portion corresponds to rejected or non-retained candidates.
(b) Distribution of the number of execution passes required to finalize canonical traces among accepted items. Together, the panels show that benchmark construction is both selective and iterative rather than one-shot.}
    \label{fig:annotation_stats}
\end{figure}

\begin{figure}[t]
    \centering
    \begin{minipage}[t]{0.73\textwidth}
        \vspace{0pt}
        \centering
        \includegraphics[width=\linewidth]{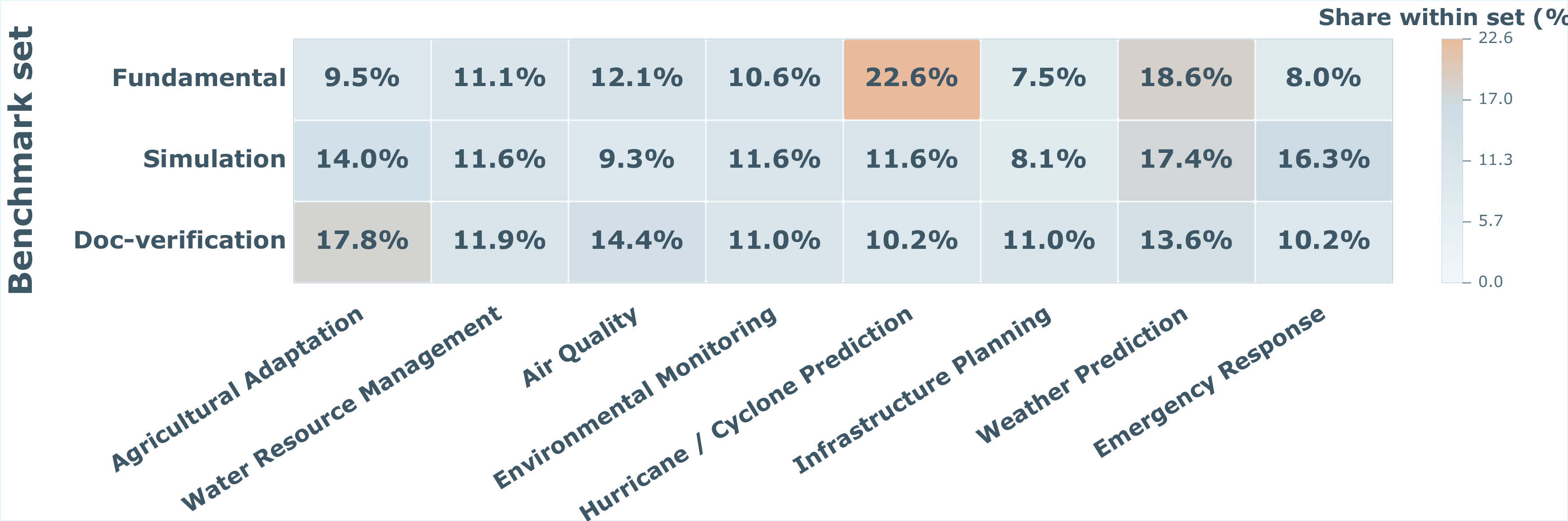}
        {\small (a)}
    \end{minipage}
    \hfill
    \begin{minipage}[t]{0.25\textwidth}
        \vspace{0pt}
        \centering
        \resizebox{\linewidth}{!}{%
        \begin{tabular}{@{}lr@{}}
            \toprule
            \textbf{Tool Group} & \textbf{Sub-tools} \\
            \midrule
            Data I/O                    & 3  \\
            Ensemble \& Verification    & 12 \\
            Forecast                    & 2  \\
            GIS / OSM                   & 6  \\
            Computation                 & 2  \\
            Reanalysis / Environmental  & 12 \\
            Satellite / EO              & 6  \\
            Seasonal                    & 13 \\
            Simulation                  & 11 \\
            Visualization               & 8  \\
            Web Search                  & 2  \\
            \midrule
            Total                       & 77 \\
            \bottomrule
        \end{tabular}
        }
        {\small (b)}
    \end{minipage}
    \caption{\textbf{Benchmark composition and tool coverage.}
    (a) Row-normalized benchmark composition across application domains; each cell reports the share of tasks within a benchmark track assigned to that domain.
    (b) Tool groups and number of sub-tools available in the TerraBench execution environment.}
    \label{fig:benchmark_composition}
\end{figure}

\begin{figure}[t]
    \centering
    \begin{minipage}[t]{\textwidth}
        \centering
        \includegraphics[width=\linewidth]{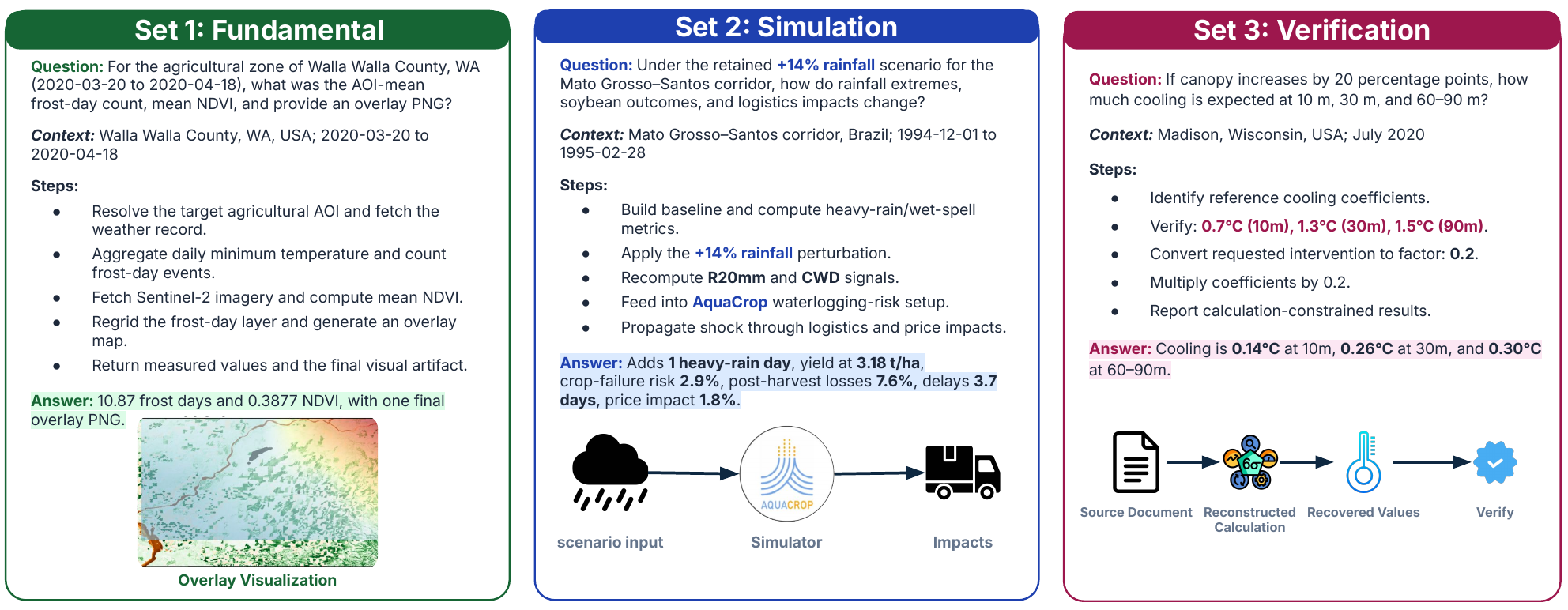}
    \end{minipage}
    \caption{Example questions from the TerraBench benchmark illustrating the three tracks.}
    \label{fig:workflow_stats}
\end{figure}

TerraBench is constructed through a semi-automated, human-in-the-loop annotation pipeline (Figure~\ref{fig:data_curate}). The final benchmark contains 403 items across three tracks and eight application domains, supported by 77 sub-tools. An partial inter-annotator agreement audit on a random subset of traces demonstrated substantial agreement (Cohen’s $\kappa = 0.694$), confirming the reliability of our manual verification protocol (see Appendix~\ref{app:annotator_agreement} for details). Figure~\ref{fig:benchmark_composition} summarizes benchmark composition and tool coverage, while Figure~\ref{fig:workflow_stats} illustrates representative tasks from the three tracks and shows that most items require multi-step, multi-tool workflows rather than simple question answering. In particular, many traces contain 50--60 steps, and most questions use 5--6 distinct tool groups, reflecting the workflow breadth required for grounded Earth-science reasoning. Unlike conventional QA examples, each item is a high-information benchmark program with contextual task framing, a structured output schema, an instruction-guided workflow, a canonical reasoning trace, tool observations, supporting artifacts, and a verified final answer. Annotation is inherently iterative rather than one-shot: frontier-capable agents assist with trace continuation, error cleanup, question normalization, and output-legitimacy checks, but final benchmark acceptance remains gated by post-execution verification and human review. Figure~\ref{fig:annotation_stats} quantifies this burden: only 50.9\% of proposed questions are retained overall, with Document-Grounded Verification being the most selective track at 38.4\%; only 28.4\% of proposed questions are retained without revision; and only 25.6\% of finalized items complete in a single run, while 74.4\% require multiple execution passes and 46.4\% require at least three runs. These statistics support our design choice to prioritize annotation depth, scientific grounding, and trace auditability over raw benchmark scale. All finalized traces are reviewed under a common protocol: a valid trace must follow the intended scientific workflow, support every retained numeric or factual claim with tool execution or artifacts. A trace is marked completed only if the final answer resolves the benchmark question and conforms to the required structured output format. For document-grounded tasks, annotators maintain a strict separation between document truth and execution truth, where reported scientific quantities serve as verification anchors rather than direct fill values. Editing is intentionally minimal, and annotators may not introduce new scientific claims, add new reasoning steps, or alter retained numbers. For numeric outputs, annotators also manually review the output to ensure that it is scientifically reasonable and neither unrealistically strict nor overly permissive. 
While TerraBench prioritizes annotation depth over raw scale, rigorous item-level bootstrap resampling ($B=2000$) confirms our sample size is statistically sufficient to yield robust, highly discriminative performance gaps (e.g., Claude Sonnet 4.6 exceeds Qwen3.5-35B by 0.1927 ToolUseScore with a 95\% CI of [0.1718, 0.2143], Appendix~\ref{app:bootstrap_results}). Overall, benchmark construction and annotation consumed approximately 2.47 billion tokens, while evaluating the frontier and baseline models consumed an additional 820 million tokens, reflecting the long-horizon complexity of the Earth-science tasks. Inference time is reported in Appendix~\ref{app:inference_time}

\subsection{Evaluation Methodology}

\begin{figure}[t]
    \centering
    \begin{minipage}[t]{0.47\textwidth}
        \centering
        \includegraphics[width=\linewidth]{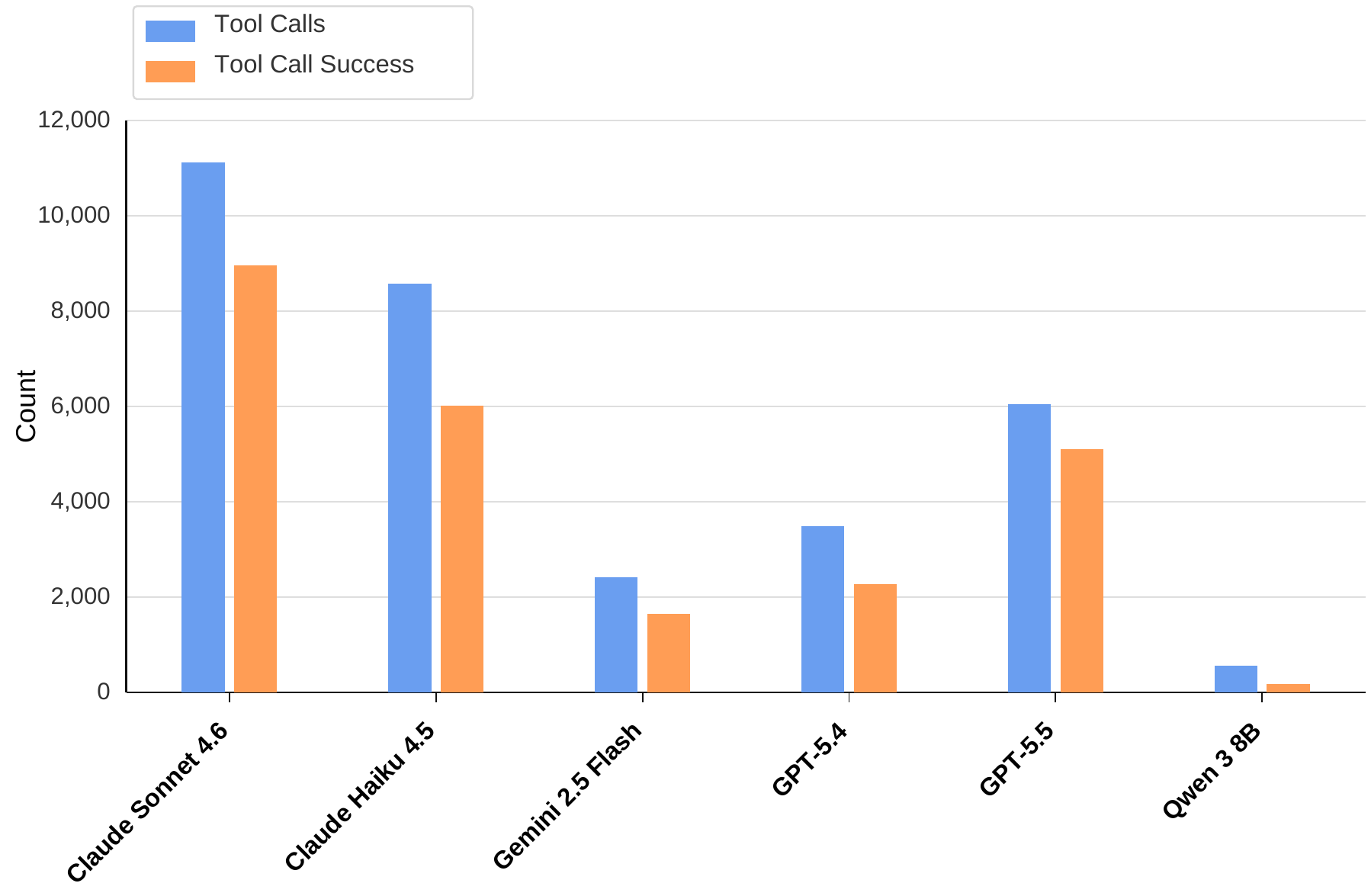}
        {\small (a)}
    \end{minipage}
    \begin{minipage}[t]{0.47\textwidth}
        \centering
        \includegraphics[width=\linewidth]{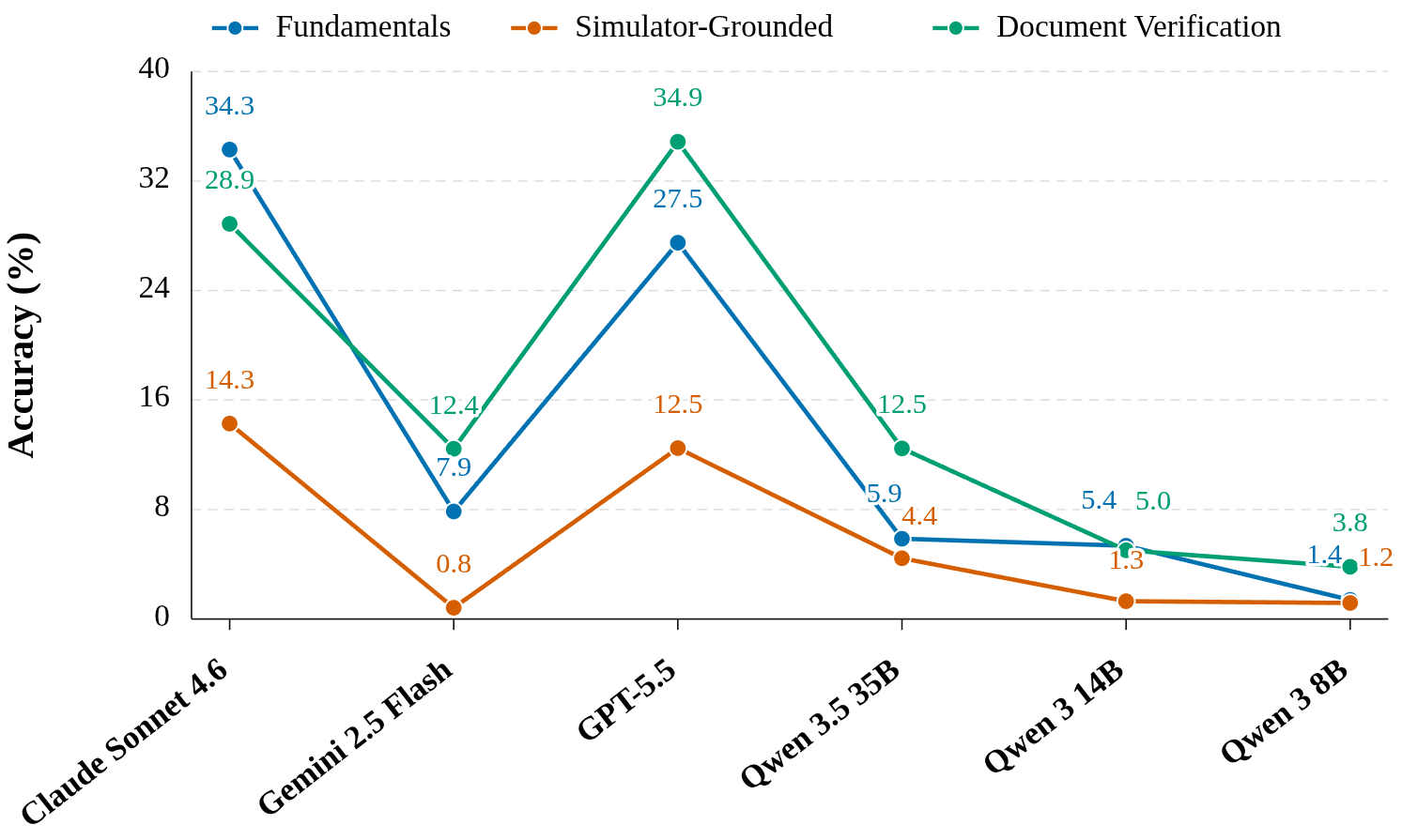}
        {\small (b)}
    \end{minipage}
    \begin{minipage}[t]{0.5\textwidth}
        \centering
        \includegraphics[width=\linewidth]{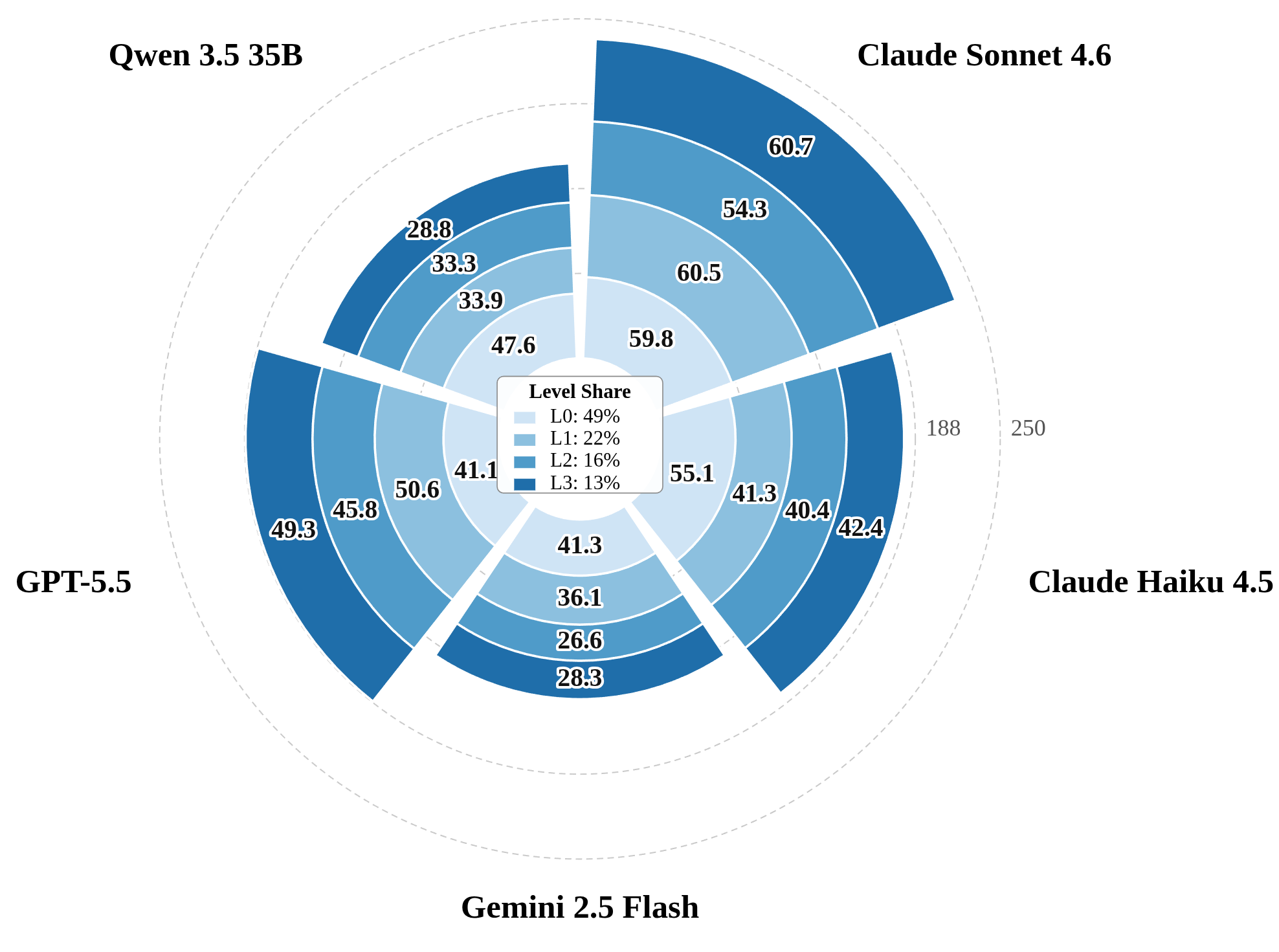}
        {\small (c)}
    \end{minipage}
    \begin{minipage}[t]{0.45\textwidth}
        \centering
        \includegraphics[width=\linewidth]{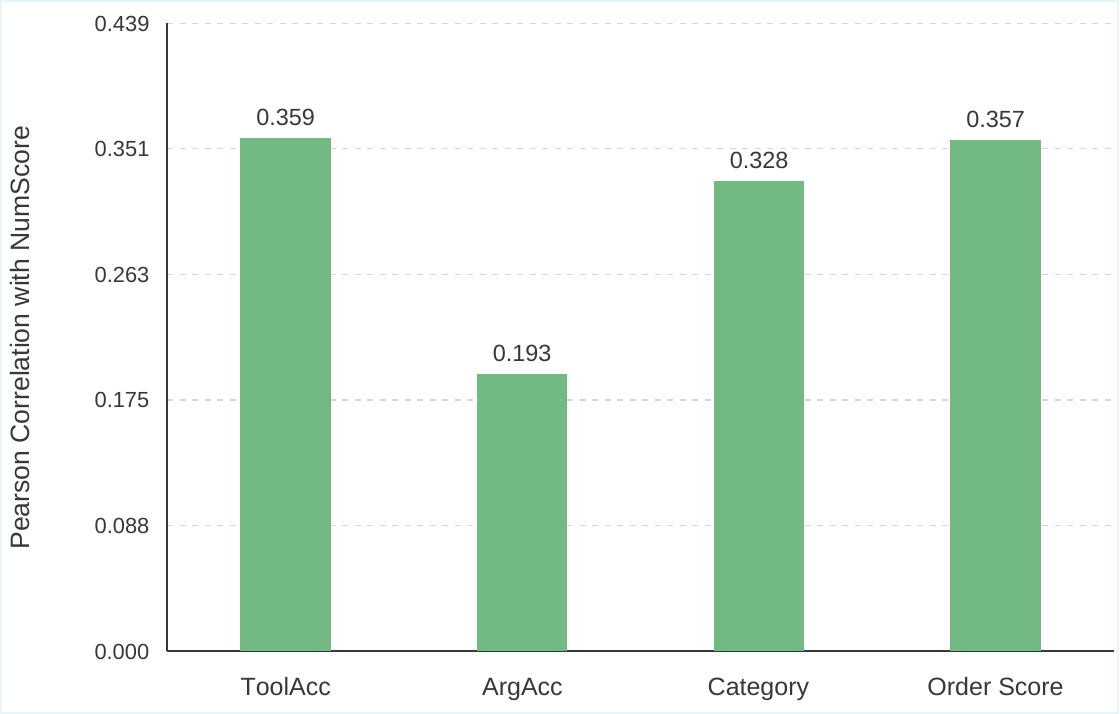}
        {\small (d)}
    \end{minipage}
    \caption{\textbf{Score analysis across models, tracks, reasoning levels, and process--outcome coupling.}
(a) Total \& successful tool calls for representative models. 
(b) NumScore by benchmark track. 
(c) Performance by reasoning level. 
(d) Pearson correlation between tool-use metrics \& NumScore.}
    \label{fig:score_analysis}
\end{figure}

\subsubsection{Tool-Use and Process Metrics}

Inspired by GTA~\cite{gta} and OpenEarthAgent~\cite{openearthagent}, we report six row-level process metrics: \textbf{InstAcc} (instruction validity), \textbf{ToolCallSuccessRate} (execution success), \textbf{ToolAcc} (tool-selection accuracy), \textbf{CategoryF1} (tool-group overlap), \textbf{ArgAcc} (argument correctness), and \textbf{OrderScore} (workflow-order consistency). These are combined into the main composite metric, \textbf{ToolUseScore}.  We assign the largest weights to tool selection and argument correctness because using the wrong scientific tool or incorrect spatiotemporal/variable parameters typically invalidates downstream Earth-science computations; category coverage, order consistency, instruction validity, and execution success provide complementary workflow diagnostics. Specific weights are provided in Appendix~\ref{app:tool_metrics} . 


\subsubsection{Answer Evaluation and Unified NumScore}

For answer evaluation, the model must return a machine-readable \texttt{<final\_json>} prediction along with free form text. The sidecar is first checked against a strict schema requiring each field to contain exactly:
\texttt{key}, \texttt{value}, \texttt{abs\_tol}, \texttt{rel\_tol}, and \texttt{floor\_scale}.

For numeric fields, let $y$ be the ground-truth value and $\hat{y}$ the prediction. We define the effective tolerance, absolute error, and normalized error as
\begin{equation}
    \lambda = \max(\tau_{\mathrm{abs}}, \tau_{\mathrm{rel}}|y|, f),
    \qquad
    AE = |\hat{y} - y|,
    \qquad
    n = \frac{AE}{\lambda}.
\end{equation}
Here $\tau_{\mathrm{abs}}$ is the absolute tolerance, $\tau_{\mathrm{rel}}$ is the relative tolerance, and $f$ is a field-specific floor scale.

We compute two complementary scores:
\begin{equation}
    \mathrm{Hit@tol}_{\mathrm{field}} = \mathbf{1}\{AE \le \lambda\},
    \qquad
    \mathrm{NumScore}_{\mathrm{field}} =
    \begin{cases}
        1, & n \le 1,\\
        2^{-(n-1)}, & n > 1.
    \end{cases}
\end{equation}
\textbf{Hit@tol} is a strict pass/fail metric, while \textbf{NumScore} provides partial credit for near misses. We use an exponential decay with base 2 so that the score halves for each additional tolerance-width of error beyond the acceptable band: predictions at $2\lambda$, $3\lambda$, and $4\lambda$ errors receive scores $1/2$, $1/4$, and $1/8$, respectively. This preserves a sharp distinction between in-tolerance and out-of-tolerance answers while avoiding an all-or-nothing penalty for small numeric deviations. For multi-field outputs, both \textbf{Hit@tol} and \textbf{NumScore} are averaged over aligned ground-truth fields. Missing or unparseable numeric predictions receive zero credit. For non-numeric fields such as strings or booleans, we use canonical exact matching, and the resulting direct-match value is used for both metrics.
\section{Experiments}

\begin{table*}[t]
\centering
\Large
\setlength{\tabcolsep}{7pt}
\renewcommand{\arraystretch}{1.12}
\resizebox{\textwidth}{!}{%
\begin{tabular}{llccccccc}
\toprule
\multirow{2}{*}{\textbf{Model}} & \multirow{2}{*}{\textbf{Param.}} &
\multicolumn{5}{c}{\textbf{Tool use}} &
\multicolumn{2}{c}{\textbf{Answer score}} \\
\cmidrule(lr){3-7} \cmidrule(lr){8-9}
& & \textbf{ToolAcc} & \textbf{CategoryF1} & \textbf{ArgAcc} & \textbf{OrderScore} & \textbf{ToolUseScore} & \textbf{NumScore} & \textbf{Hit@tol} \\
\midrule

\rowcolor{sectiongray}
\multicolumn{9}{c}{\textsc{Frontier Models}} \\

GPT-5.4  & -- & 23.88 & 32.94  & \underline{54.10} & 33.02  & 41.09  & 12.29 & 9.80 \\
GPT-5.5  & -- & 35.39  & 37.22 & 39.73 & \underline{39.67} & \underline{45.01} & \underline{26.45} & \underline{21.03}  \\
Gemini 3.1 Pro Preview & -- & 35.64 & 32.11 & 46.91 & 42.07 & 44.93 & 17.07 & 13.21 \\
Gemini 2.5 Flash & -- & 18.46 & 24.16 & 50.91 & 23.97 & 36.20 & 7.70 & 6.34 \\
Claude Haiku 4.5 & -- & \underline{38.25}  & \underline{40.26}  &  51.25 & 46.79 & 48.14 & 21.19 & 15.23 \\
Claude Sonnet 4.6 & -- & \textbf{54.43}  & \textbf{50.05} & \textbf{52.26} & \textbf{61.02} & \textbf{59.22} & \textbf{28.44}  & \textbf{22.88}  \\

\midrule
\rowcolor{sectiongray}
\multicolumn{9}{c}{\textsc{Baseline Agent}} \\
Qwen3.5-9B  & 9B & 21.56 & 23.57 & 37.53 & 25.62 & 31.18 & 1.30 & 1.20 \\
\midrule
\rowcolor{sectiongray}
\multicolumn{9}{c}{\textsc{Open-Source Models}} \\
Qwen3-1.7B     & 1.7B  & 1.02 & 1.85 & 7.03 & 1.74 & 5.40  & 1.15 & 1.08 \\
Gemma 4 E4B              & 4B  & 5.90 & 9.00 & 29.05 & 8.50 & 16.31 & 3.04 & 2.90 \\
Mistral 7B Instruct v0.3 & 7B  & 6.40 & 10.99 & 37.19 & 10.01 & 18.79 & 2.41 & 1.67 \\
Llama 3.1 8B  & 8B  & 0.20 & 0.40 & 1.80 & 0.40 & 1.50 & 1.00 & 0.90 \\
InternVL3-8B             & 8B & 1.60 & 2.70 & 7.50 & 2.30 & 4.60 & 1.70 & 1.10   \\
Qwen3-8B     & 8B  & 2.42 & 4.07 & 14.90 & 3.77 & 7.97 & 2.06 & 1.33 \\
Qwen3-VL-8B  & 8B  & 7.67 & 12.17 & 33.09 & 11.35 & 19.09 & 5.07 & 3.11 \\    
Qwen3-14B    & 14B  & 9.40 & 13.66 & 37.72 & 13.32 & 21.12 & 4.30 & 3.80 \\
Gemma 4 26B-A4B               & 26B & \textbf{31.89} & 28.74 & 43.69 & 35.05 & \textbf{41.84} & 4.45 & 3.67 \\
Qwen3.5-35B-A3B        & 35B & 30.17 & \textbf{30.05} & \textbf{47.23} & \textbf{36.46} & 39.95 & \textbf{7.49} & \textbf{5.89} \\
\bottomrule
\end{tabular}%
}
\caption{Main quantitative results on TerraBench. Models are evaluated across fine-grained tool-use process metrics and strict answer-correctness metrics. The best overall score in each column is \textbf{bolded}, and the second-best is \underline{underlined}.}
\label{tab:main_results}
\end{table*}

\subsection{Quantitative Analysis}

Table~\ref{tab:main_results} shows that TerraBench is challenging for both frontier and open models. Even the strongest reported model, Claude Sonnet 4.6, reaches only 59.22 ToolUseScore, 28.44 NumScore, and 22.88 Hit@tol, leaving substantial headroom on both process and outcome dimensions. 
Among other frontier models, performance diverges between process and outcome: Claude Haiku 4.5 achieves the second-highest process quality (48.14 ToolUseScore), while GPT-5.5 secures the second-best final outcomes (26.45 NumScore and 21.03 Hit@tol). By contrast, open-weight models show a distinct capability split: Gemma 4 26B-A4B demonstrates the strongest workflow execution (41.84 ToolUseScore), whereas Qwen3.5-35B-A3B yields the best outcome metrics (7.49 NumScore and 5.89 Hit@tol). This indicates a substantial frontier--open gap on both workflow execution and final answer grounding. A consistent pattern across models is that process quality is higher than answer quality. ToolUseScore is generally much higher than both NumScore and Hit@tol, suggesting that partial tool orchestration is easier than producing fully grounded, tolerance-satisfying outputs. This is especially visible for compact and mid-sized open models: for example, Qwen3.5-9B achieves a non-trivial 31.18 ToolUseScore, but only 1.30 NumScore and 1.20 Hit@tol. This makes Qwen3.5-9B a useful lower-bound agentic baseline: it is small enough to be reproducible and affordable, yet strong enough to interact meaningfully with the benchmark's tool environment. At the same time, the gap between Qwen3.5-9B and stronger frontier models shows that basic tool access alone is insufficient for robust climate reasoning. These results suggest a two-stage capability gap. \textit{First}, models must learn to select, order, and execute the right tools. \textit{Second}, and more difficult, they must use those tools precisely enough to produce structured outputs that remain numerically correct and scientifically grounded across EO, environmental, GIS, and simulator-backed workflows. Track-level analysis further indicates that the benchmark is not homogeneous: Fundamentals is generally the easiest setting, Simulator-Grounded tasks are the hardest overall, and Document-Grounded Verification probes a distinct capability mixture rather than simply behaving as a more difficult version of the other two tracks. Overall, TerraBench is strongly discriminative across model scales and exposes clear limitations in current frontier and open LLM agents beyond shallow question answering.

\subsection{Tool use vs. Answer correlation}

\begin{figure}[t]
    \centering
    \begin{minipage}[t]{0.47\textwidth}
        \centering
        \includegraphics[width=\linewidth]{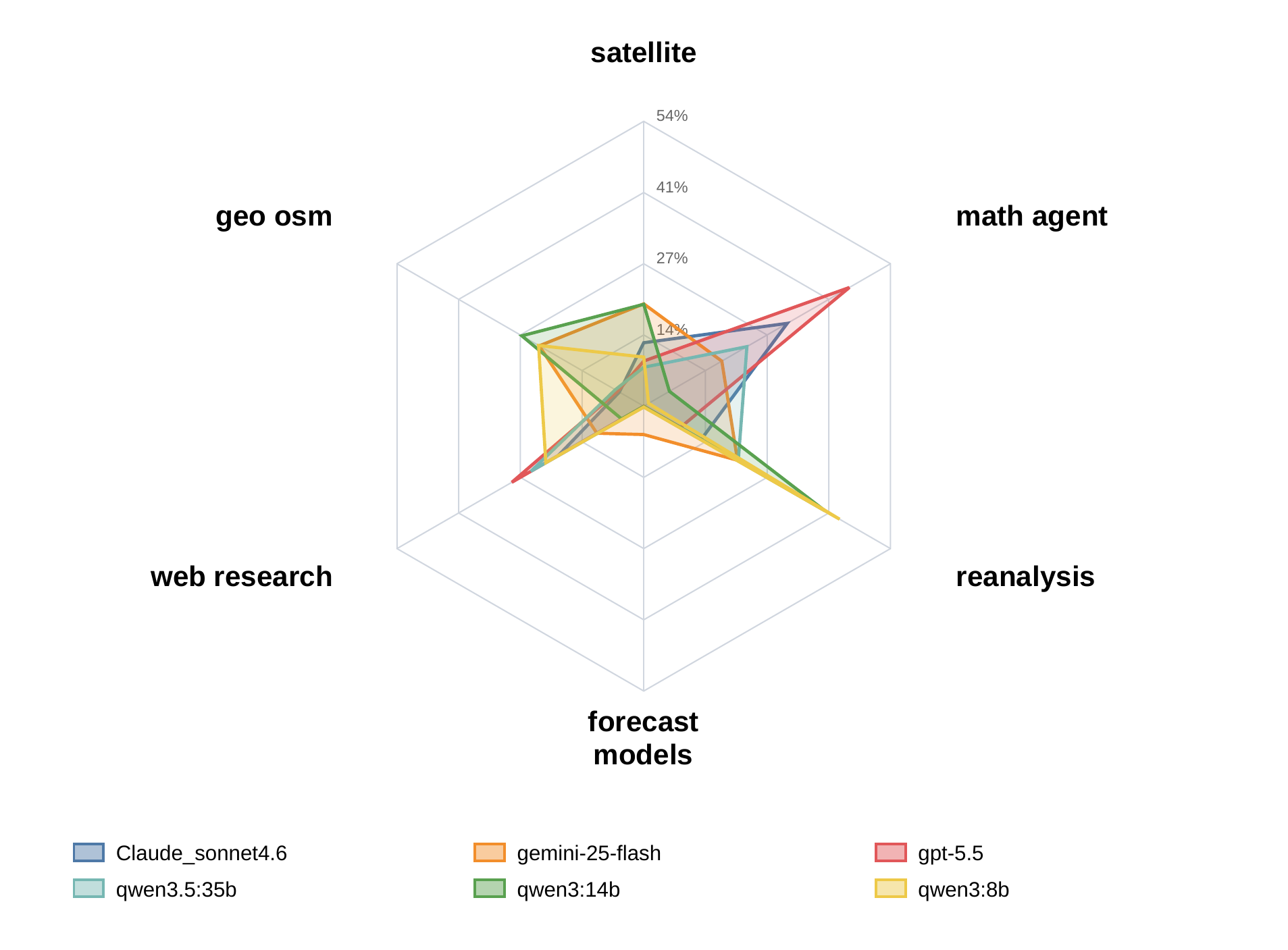}
        {\small (a)}
    \end{minipage}
    \begin{minipage}[t]{0.52\textwidth}
        \centering
        \includegraphics[width=\linewidth]{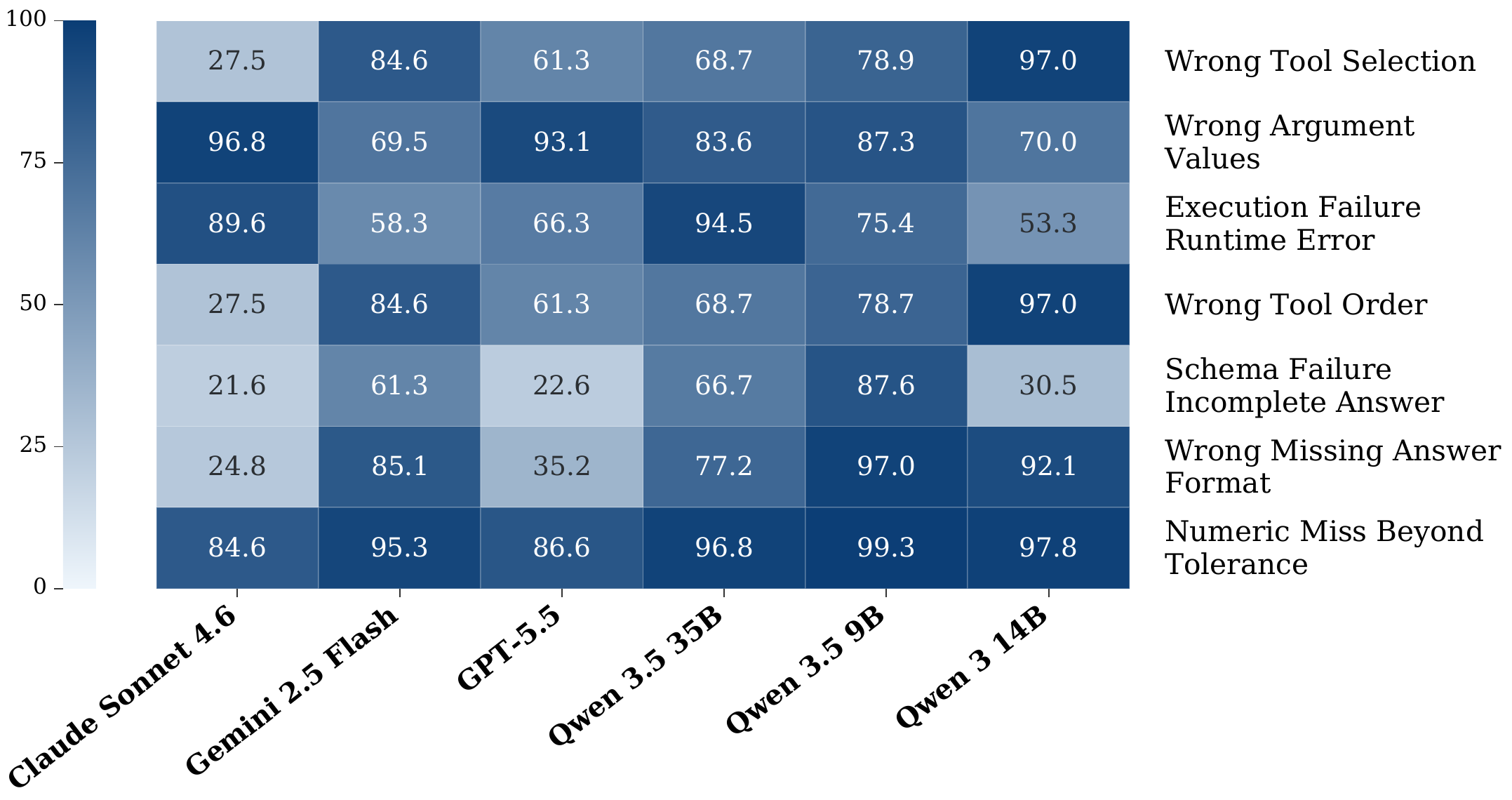}
        {\small (b)}
    \end{minipage}
    \caption{\textbf{Tool-family usage and dominant failure categories.}
(a) Relative distribution of tool-family usage across representative models. 
(b) Share of failed items exhibiting each dominant failure mode, including tool-selection, argument, execution, ordering, schema, answer-format, and numeric-grounding failures.}
    \label{fig:tool_error_analysis}
\end{figure}

Figure~\ref{fig:score_analysis}(d) measures the Pearson correlation between core tool-use metrics and the final numeric-answer metric, NumScore. All four correlations are positive but moderate: ToolAcc ($r=0.359$), OrderScore ($r=0.357$), CategoryF1 ($r=0.328$), and ArgAcc ($r=0.193$). This indicates that stronger tool orchestration is associated with better numeric outcomes, but does not by itself determine them. In particular, selecting the right tools and ordering them coherently appear more predictive of final correctness than argument accuracy alone. Therefore, a model can partially follow the correct workflow yet still fail numerically, and conversely may obtain a partially correct answer despite imperfect tool use. This is also consistent with the broader results in Table~\ref{tab:main_results}, where process metrics are generally higher than answer metrics across models.

\subsection{Failure-mode decomposition}

Figure~\ref{fig:tool_error_analysis} shows that the dominant bottlenecks in TerraBench lie in argument grounding, workflow orchestration, and final numeric grounding rather than in any single isolated tool family. Across all currently analyzed models, Numeric Miss Beyond Tolerance remains the most universal failure mode, ranging from 84.6\% for Claude Sonnet 4.6 to 99.3\% for Qwen3.5-9B, while Wrong Argument Values is also consistently high, spanning 69.5\%--96.8\%. Tool-orchestration failures remain widespread, but with substantial variation across models: Claude Sonnet 4.6 shows much lower rates of Wrong Tool Selection and Wrong Tool Order (27.5\% each) than weaker models such as Qwen3:14B (97.0\% each), suggesting that frontier gains come partly from better workflow planning rather than from answer formatting alone. At the same time, answer-side failures remain severe for several models, with Wrong / Missing Answer Format reaching 85.1\% for Gemini 2.5 Flash, 77.2\% for Qwen3.5-35B, 97.0\% for Qwen3.5-9B, and 92.1\% for Qwen3:14B. Finally, Figure~\ref{fig:tool_error_analysis}(a) shows that different models rely on different subsets of the available tool environment. Stronger models make broader use of reanalysis, GIS/OSM, forecast, satellite, and math-agent utilities, whereas weaker models under-utilize the benchmark’s tool diversity and collapse onto a narrower subset of actions. Simulation and visualization tools are rarely used at all. For example, Claude Sonnet 4.6 allocates only about 1.31\% of tool calls to visualization, and GPT-5.5 uses simulation tools in only about 0.28\% of calls. This indicates that many models optimize for producing a direct textual answer rather than externalizing intermediate evidence through maps, plots, or simulator-backed analyses, which likely contributes to the particularly weak performance observed on Simulator-Grounded tasks.

\section{Conclusion}

We introduced TerraBench and TerraAgent, a benchmark and executable framework for tool-augmented Earth-science reasoning. TerraBench evaluates LLM agents on coordinating multimodal inputs, simulators, and document-grounded evidence to produce structured numerical outputs. While frontier models outperform open baselines on process metrics, all exhibit a significant gap between tool-use proficiency and final-answer correctness. TerraBench and TerraAgent provide a reproducible testbed to guide the development of scientifically grounded LLM agents for Earth-science.

\textbf{Limitations:}
TerraBench is designed to prioritize annotation depth and scientific rigor, which naturally shapes two aspects of the current release. First, benchmark quality is ensured through strict post-execution verification tied to deterministic workflows, where classical inter-annotator agreement on free-form traces is less meaningful than auditing benchmark-critical decisions such as numeric tolerance validation. A stratified independent expert audit is planned for future releases to further strengthen these boundaries. Second, the depth-first annotation makes each item resource-intensive to construct, reflecting a deliberate trade-off between scale and quality. Future work will explore semi-automated pipelines to reduce overhead while preserving expert-guided scientific grounding.

\begin{ack}
\paragraph{Funding disclosure.}
This material is based on work supported by ADIA Lab, Abu Dhabi under a collaborative research project titled Climate Co-Pilot: A Multimodal AI Agent for Climate Resilience in the MENA Region.

\paragraph{Competing interests.}
The authors declare no competing interests.
\end{ack}

\bibliographystyle{plain}
\bibliography{references}

\appendix

\input{Appendix}
\clearpage





\end{document}

%% file: Appendix.tex
\appendix
\clearpage

\section*{Supplementary Material\\``TerraBench: Can Agents Reason Over Heterogeneous Earth-System Data?''}
\addcontentsline{toc}{section}{Supplementary Material}
\label{app:supplement_title}

\noindent
This supplement provides additional details on the TerraAgent framework (Appendix \ref{app:framework_details}), benchmark construction (Appendix \ref{app:trace_generation}), annotation criteria (Appendix \ref{app:annotation_protocol}), evaluation metrics (Appendix \ref{app:evaluation_metrics}), bootstrap uncertainty estimates (Appendix \ref{app:bootstrap_results}), extended failure analysis (Appendix \ref{app:error_analysis}), computational resources (Appendix \ref{app:computational_resources}), experimental setup (Appendix \ref{app:experimental_setup}) and simulator descriptions (Appendix \ref{app:simulator}).

\section{TerraAgent Framework Details}
\label{app:framework_details}

TerraAgent is a tool-augmented framework for climate and environmental analysis. It couples LLM planning with tool-grounded scientific workflows so that quantitative outputs are derived from data retrieval, geospatial processing, model execution, simulation, visualization, or explicit computation rather than unsupported language-model generation.

The framework follows a \emph{ReAct}-style design \cite{react,agentscope}: for each task, the agent decomposes the request, selects an appropriate tool, observes the returned result, and iteratively refines its plan before producing a final answer. This separation between linguistic planning and scientific execution is central to the framework, ensuring that quantitative outputs are derived from data retrieval, model execution, simulation, or explicit computation rather than from unsupported language-model generation. In practice, we instantiate an AgentScope-based ReAct agent \cite{agentscope} with bounded iterative execution, serial tool use, and explicit support for trace continuation.

A key feature of TerraAgent is its \emph{domain-organized tool registry}. A small bootstrap layer remains available throughout execution, including region resolution, web search and summarization, streaming to a remote math agent, and math-agent execution. On top of this bootstrap layer, specialist tools are organized by scientific function and activated as needed. As shown in Figure~\ref{fig:framework_overview}, the registry is grouped into eight high-level tool families:
\textbf{Reanalysis / Environmental}, which retrieves and preprocesses gridded physical datasets such as ERA5 and air-quality products;
\textbf{Forecast / Seasonal}, which supports numerical forecast wrappers, C3S seasonal products, and forecast verification workflows;
\textbf{Satellite imagery}, which provides Earth observation retrieval and index computation through Sentinel-2 and Google Earth Engine;
\textbf{GIS / OSM}, which handles geospatial querying, region resolution, vector/raster operations, and OpenStreetMap features;
\textbf{Simulation}, which exposes deterministic impact and sectoral simulators such as AquaCrop~\cite{aquacrop}, DSSAT~\cite{dssat}, CLIMADA~\cite{climada}, EnergyPlus~\cite{energyplus}, UTCI/health-risk tools~\cite{utci}, and SUMO traffic simulation~\cite{sumo};
\textbf{Visualization}, which renders maps, overlays, and derived visual products;
\textbf{Web Search}, which retrieves external scientific context or methodological references when permitted; and
\textbf{Execution server}, inspired by ZephyrusWorld~\cite{varambally2025zephyrus}, which provides a Python execution service for auxiliary computations not directly supported by the main tool registry, particularly mathematical operations and lightweight scientific calculations needed to complete the reasoning workflow.

The framework is also \emph{artifact-centered}. Tools return structured objects containing both human-readable summaries and machine-readable metadata, while intermediate and final products are materialized as domain-appropriate artifacts such as NetCDF, CSV, JSON, GeoTIFF, and PNG files. Forecast wrappers normalize outputs onto common latitude--longitude grids and export them in interoperable formats, enabling downstream regridding, visualization, ensemble analysis, and verification. Similarly, simulator tools produce deterministic, provenance-bearing outputs that support direct comparison across baseline, intervention, and counterfactual scenarios.

Overall, TerraAgent operationalizes climate reasoning as an executable sequence of tool-mediated steps. By combining model-based planning, domain-specialized scientific tools, and artifact-backed computation, it provides a reproducible foundation for climate, weather, air-quality, geospatial, remote-sensing, and impact-assessment tasks.

\subsection{Tool Registry and Artifact-Centered Execution}
\label{app:tool_registry}

The framework exposes a domain-organized tool registry covering reanalysis/environmental data, forecast and seasonal products, satellite/EO retrieval and indices, GIS/OSM operations, deterministic simulation, visualization, web search, and an execution server for auxiliary scientific computation. Tools return human-readable summaries and machine-readable metadata, while intermediate and final products are materialized as domain-appropriate artifacts such as NetCDF, CSV, JSON, GeoTIFF, and PNG files. This artifact-centered design allows benchmark traces to be audited after execution and enables direct comparison across baseline, intervention, and counterfactual scenarios.

\section{Trace Generation and Benchmark Composition}
\label{app:trace_generation}

A key component of the construction pipeline is the instruction prompt, which acts as an executable workflow template during annotation and trace generation. Rather than serving as a simple answer key, the prompt constrains the annotating agent to follow a scientific workflow such as document-anchor identification, AOI/time definition, multimodal data retrieval, fusion, computation or simulation, visualization, validation, and final packaging. During annotation, the execution agent uses real tools to produce a canonical reasoning trace (\texttt{Main\_trace.json}), tool observations, supporting artifacts, and a structured ground-truth answer file (e.g., \texttt{number\_ground\_truth.json}).

\subsection{Dataset Composition by Track and Use Case}
\label{app:dataset_composition}


We organize TerraBench into three benchmark tracks: \textbf{Fundamentals}, \textbf{Simulator-Grounded}, and \textbf{Document-Grounded Verification}. All three tracks share the same visible task interface i.e., a contextualized natural-language \textit{question + context} paired with a structured output schema, but differ in how correctness is established: through deterministic execution, deterministic simulation, or quantitative anchors from real scientific documents. Across the dataset, benchmark construction required roughly \textbf{117 person-days} of expert annotation and verification. We report this statistic as a benchmark-construction metadata rather than as a direct measure of benchmark quality.

The three tracks are defined as follows:
\begin{itemize}
    \item \textbf{Fundamentals:} easy-to-verify multimodal execution tasks that combine gridded environmental data, satellite imagery or indices, and GIS features.
    \item \textbf{Simulator-Grounded:} longer-horizon agentic tasks for planning, intervention, and counterfactual analysis using deterministic simulation workflows (e.g., DSSAT, CLIMADA).
    \item \textbf{Document-Grounded Verification:} tasks that reconstruct or approximate reported scientific quantities using public data, tools, and method anchors from authoritative papers or reports.
\end{itemize}

Each retained item follows a standardized schema:
\textit{Context, Use case, Subtask, Level, Question, Reference, Answer, and Instruction Prompt}.
The visible \textit{Answer} field defines the output contract, typically via an output specification and a required \texttt{<final\_json>} structure. To prevent evaluation leakage, models at inference time are given only the visible task specification (Context, Question, scientific Reference framing, and structured output schema). Hidden benchmark labels, document anchors, and evaluator-side metadata remain strictly hidden.

Although TerraBench contains \textbf{403 benchmark items}, each item is substantially richer than a conventional QA example: beyond a contextualized question, it includes a structured output schema, an executable workflow template, a canonical reasoning trace, tool observations, supporting artifacts, and a verified final answer. The resulting benchmark should therefore be understood as a collection of \emph{high-information executable benchmark programs}, rather than static question-answer pairs.

\paragraph{Reasoning levels.}
In addition to the three tracks, we organize items into four reasoning levels (\textbf{Level 0--3}) that reflect the depth of reasoning, workflow complexity, and degree of scientific decision-making required.

\begin{itemize}
    \item \textbf{Level 0:} direct execution tasks with short reasoning chains, typically requiring retrieval, aggregation, thresholding, or a simple multimodal overlay with easy-to-verify outputs.
    \item \textbf{Level 1:} multi-step analytic tasks that require integrating multiple data products or modalities and performing structured comparison, anomaly analysis, or evidence synthesis.
    \item \textbf{Level 2:} intervention-oriented tasks that require explicit scenario setup, model or simulator invocation, and comparison between baseline and modified conditions.
    \item \textbf{Level 3:} the most demanding tasks, involving counterfactual or long-horizon reasoning, complex simulation workflows, or document-grounded reconstruction under incomplete public observability.
\end{itemize}

This level taxonomy is used to organize benchmark reasoning independently of the three primary tracks, allowing the benchmark to separate \emph{task type} from \emph{reasoning depth}.

\subsection{Dynamic Ground Truth and Verification Modes}
\label{app:ground_truth}

Following annotation, traces pass through a post-execution verification layer that audits workflow compliance, removes invalid branches, and checks the provenance of all retained outputs. Ground truth is constructed according to task type: Fundamentals use deterministic public-data workflows over environmental and satellite inputs; Simulator-Grounded tasks use deterministic simulator outputs under explicit baseline, intervention, or counterfactual assumptions; and Document-Grounded Verification tasks use a dual-layer design consisting of document truth (the reported number and source location) and execution truth (the benchmark reconstruction using public tools). To preserve scientific realism, the document-grounded track uses explicit verification modes: exact reproduction, approximate reproduction, structural verification, and proxy verification. Near matches and scientifically reasonable approximations are accepted only when the construct, units, and computational setup align with the source document.

\section{Human Annotation Criteria and Review Protocol}
\label{app:annotation_protocol}

\subsection{Benchmark Creation Pipeline}
\label{sec:benchmark_creation}

We construct TerraBench through a hybrid generation, annotation, and verification pipeline designed to produce executable, scientifically grounded benchmark items rather than static question--answer pairs. The benchmark is organized into three tracks---\textbf{Fundamentals}, \textbf{Simulator-Grounded}, and \textbf{Document-Grounded Verification}---covering \textbf{8 application domains} and supported by an executable environment with \textbf{77 sub-tools}. All three tracks share the same user-facing task interface: a contextualized natural-language \textit{question + context} paired with a structured output schema, even though the source of ground truth differs across tracks.

Our pipeline is inspired by recent agentic benchmark construction frameworks that combine structured task authoring, executable workflows, and post-hoc verification \cite{agentx,varambally2025zephyrus}. The process begins with broad task ideation. We first brainstorm domain-relevant use cases across practical climate and environmental applications, and then use LLM-assisted web research and document extraction to identify candidate scientific workflows, operational scenarios, and document-grounded quantitative results. This produces an initial pool of candidate tasks spanning all three tracks. In the current benchmark release, this initial pool contains \textbf{792 candidate questions}: \textbf{323 Fundamentals}, \textbf{307 Document-Grounded Verification}, and \textbf{162 Simulator-Grounded} candidates. These are subsequently filtered through feasibility checks, execution-based annotation, multi-stage verification, and rejection or major revision, yielding a final benchmark of \textbf{403 validated items}.

Each retained item is authored with a standardized schema:
\textit{Context, Use case, Subtask, Level, Question, Reference, Answer, and Instruction prompt}.
The visible \textit{Answer} field defines the output contract, typically via an output specification and a required \texttt{<final\_json>} structure. Hidden benchmark metadata, document anchors, and evaluator-side labels are never exposed at inference time, which keeps the model interface consistent across tracks while preventing benchmark leakage.

To support reproducible trace generation, we design a family of \emph{workflow templates} that capture the reasoning structures we expect the annotating agent to follow, including reanalysis + Earth observation fusion, multi-model ensemble reasoning, simulation execution, and structured result packaging. While these templates share a common high-level scaffold, they are specialized for each track to reflect track-specific requirements.

A key component of the pipeline is the \textbf{Instruction Prompt}, which functions as an executable workflow template rather than a simple answer key. Without this structured prompt, autonomous agents often skip required workflow stages, omit validation checks, or drift toward unsupported computational shortcuts. The instruction prompt therefore enforces a canonical reasoning structure over AOI/time setup, multimodal data retrieval, fusion, computation or simulation, validation, visualization, and final answer packaging. During annotation, an execution agent follows this template using real tools to generate an initial reasoning trace, supporting artifacts, and a candidate final answer.

Ground truth is constructed differently for each track. For \textbf{Fundamentals}, ground truth is derived directly from deterministic public-data workflows over environmental grids, satellite imagery, and GIS layers. For \textbf{Simulator-Grounded} tasks, ground truth is defined by deterministic simulator outputs under explicit baseline, intervention, or counterfactual assumptions. For \textbf{Document-Grounded Verification}, the benchmark uses a dual-layer ground-truth design: \emph{document truth} (the reported quantity and its source location) and \emph{execution truth} (the benchmark's public-data reconstruction of the same construct or a scientifically justified analogue). This is what grounds the third track in real scientific results while preserving the same visible input interface as the other two.

To maintain scientific realism, the document-grounded track uses explicit verification modes---\textit{exact reproduction}, \textit{approximate reproduction}, \textit{structural verification}, and \textit{proxy verification}---rather than requiring strict equality in all cases. A soft agreement policy allows near matches and scientifically justified approximations when the construct, units, and computational setup align with the source document. This ensures that the benchmark evaluates grounded recovery of real scientific quantities rather than pure document extraction.

Following annotation, each trace passes through a \textbf{post-execution verification layer}. This stage audits workflow compliance, removes invalid or abandoned branches, checks provenance of retained outputs, confirms that required fields are supported by the surviving trace, and canonicalizes a single final trace for evaluation. The final artifacts are packaged into parallel prediction and ground-truth folder structures centered around canonical \texttt{Main\_trace.json} files and structured ground-truth JSONs such as \texttt{number\_ground\_truth.json}.

Overall, TerraBench should be understood not as a static QA collection, but as a benchmark of \textbf{high-information executable benchmark programs}. Each item includes multimodal inputs, a structured output contract, an executable workflow template, a canonical reasoning trace, tool observations, and verified outputs. The benchmark creation pipeline can therefore be summarized as in figure

\subsection{Review protocol}
\label{app:Review protocol}

All benchmark traces are annotated under a common set of criteria that apply equally to human annotators and to any AI support tools used during the annotation process. In our pipeline, AI assistance is used only to accelerate routine operations such as error identification, structural cleanup, trace continuation, and retrieval of missing field-specific background knowledge. Final benchmark decisions remain human-controlled: annotators are responsible for scientific judgment, workflow correction, benchmark acceptance, and final approval of the canonical trace.

We have the following annotation criteria throughout the pipeline.

\paragraph{(1) Workflow fidelity.}
A valid trace must follow the intended scientific workflow for its task type. Annotators check that the retained reasoning path respects the required workflow family specified by the instruction prompt, including task setup, AOI/time definition, multimodal data retrieval, fusion, computation or simulation, validation, and final answer packaging. Traces that bypass required stages through unsupported shortcuts are not accepted as completed artifacts.

\paragraph{(2) Provenance and computational support.}
Every retained final numeric or factual claim must be supported by surviving tool observations, retained artifacts, or transparent deterministic transforms of such outputs. Annotators do not accept values that are fabricated, manually inserted without evidence, or copied from hidden benchmark metadata. The goal is to ensure that each finalized trace remains an auditable execution record rather than a post-hoc summary.

\paragraph{(3) Leakage control.}
Annotators remove benchmark-only scaffolding that should not appear in the final trace, including annotation-only instruction text, hidden answer cues, evaluator-side labels, and unnecessary reference leakage. In particular, document-grounded values and other hidden benchmark metadata are not allowed to directly populate computed output fields. This preserves a strict separation between visible task content and hidden benchmark truth.

\paragraph{(4) Branch management and trace honesty.}
The finalized trace should preserve the real chronological reasoning path. Dead branches, exact duplicates, and abandoned retries may be removed when they have no downstream value. However, failures or partial attempts are retained when they meaningfully explain a later pivot, fallback, or proxy choice. 

\paragraph{(5) Question-answer completeness.}
A trace is marked \texttt{completed} only if the surviving final answer actually answers the benchmark question. For multi-part questions, all requested outputs, comparisons, thresholds, or decisions must be addressed. Supported numeric values alone are not sufficient: the final answer must unambiguously map those values to the quantities requested in the question. If the trace is well supported but still fails to answer the question completely, it is retained only as a partial artifact and is not classified as a completed benchmark instance and it will move to revision stage. 

\paragraph{(6) Output-schema compliance.}
Annotators verify that the final answer conforms to the visible output schema, including the required \texttt{<final\_json>} structure, canonical field names, and expected units or sign conventions.

\paragraph{(7) Document-grounded verification discipline.}
For document-grounded tasks, annotators keep a strict distinction between \emph{document truth} and \emph{execution truth}. Reported scientific quantities are treated as anchors for verification, not as direct fill values for computed outputs. Traces are accepted when the reconstructed quantity is scientifically aligned with the source document, but annotators are not allowed to force equality by injecting the reported target into the trace.

\paragraph{(8) Minimal and coherent editing.}
Human annotators may clean wording, remove irrelevant path, and rewrite assistant thought blocks minimally so that the surviving trace remains coherent and readable. However, they may not add new reasoning steps, introduce new scientific claims, or alter retained numbers. The final artifact should read as a cleaned version of the real execution trace, not as a newly authored solution.

In summary, the human annotation protocol treats traces as \emph{scientific execution artifacts} rather than free-form reasoning logs. AI assistance is used only to speed up routine maintenance and fill domain knowledge gaps; all benchmark-critical decisions---including whether a trace is valid, complete, and scientifically acceptable---remain under human control.
\section{Formal Evaluation Metrics}
\label{app:evaluation_metrics}

TerraBench evaluates models along two complementary dimensions: process quality and answer quality. Process evaluation measures whether the model selects appropriate tools, supplies correct arguments, orders steps coherently, and executes calls successfully. Answer evaluation measures whether the final structured output is schema-valid, numerically grounded, and semantically aligned with the target. Evaluation is performed over paired benchmark items, and aggregate metrics are unweighted row means so that the benchmark item, rather than the tool step or output field, is the unit of evaluation.

\subsection{Tool-Use and Process Metrics}
\label{app:tool_metrics}

Tool steps are extracted from prediction and ground-truth traces and aligned with dynamic programming under exact tool-name matching or relaxed matching at the tool-group level. Let $P_{\mathrm{all}}$ denote all predicted non-meta tool calls, $P_{\mathrm{valid}}$ the subset of valid predicted calls, $P_{\mathrm{success}}$ the subset of predicted calls with successful observations, and $G_{\mathrm{valid}}$ the valid ground-truth tool calls.

\paragraph{Instruction Accuracy.}
\begin{equation}
    \mathrm{InstAcc} = \frac{|P_{\mathrm{valid}}|}{|P_{\mathrm{all}}|}.
\end{equation}

\paragraph{Tool Call Success Rate.}
\begin{equation}
    \mathrm{ToolCallSuccessRate} = \frac{|P_{\mathrm{success}}|}{|P_{\mathrm{all}}|}.
\end{equation}

\paragraph{Tool Accuracy.}
After alignment, relaxed tool accuracy is the fraction of valid ground-truth steps matched either by exact tool identity or by the same metric/tool group:
\begin{equation}
    \mathrm{ToolAcc} = \frac{\mathrm{relaxed\_matches}}{|G_{\mathrm{valid}}|}.
\end{equation}

\paragraph{CategoryF1 and ArgAcc.}
CategoryF1 is a multiset F1 score over tool-group labels. Argument correctness (ArgAcc) averages semantic key matching and value matching over aligned tool pairs. It captures errors in temporal windows, thresholds, bounding boxes, selected variables, and other tool parameters.

\paragraph{OrderScore.}
OrderScore compares predicted and reference tool-group sequences:
\begin{equation}
    \mathrm{OrderScore} = \frac{\mathrm{Unique} + \mathrm{AnyOrder} + \mathrm{SameOrder}}{3},
\end{equation}
where \texttt{Unique} measures set overlap over tool groups, \texttt{AnyOrder} measures multiset overlap, and \texttt{SameOrder} measures longest-common-subsequence agreement.

\paragraph{Composite ToolUseScore.}
\begin{equation}
\begin{aligned}
\mathrm{ToolUseScore} =\;&
0.30 \cdot \mathrm{ToolAcc}
+ 0.15 \cdot \mathrm{InstAcc}
+ 0.20 \cdot \mathrm{ArgAcc} \\
&+ 0.15 \cdot \mathrm{CategoryF1}
+ 0.15 \cdot \mathrm{OrderScore}
+ 0.05 \cdot \mathrm{ToolCallSuccessRate}.
\end{aligned}
\end{equation}

\paragraph{Motivation.}
\textsc{ToolUseScore} is an a priori composite process metric designed to summarize whether an agent follows a scientifically valid tool-use workflow. It combines six process components: tool accuracy, instruction validity, argument correctness, category-level tool coverage, workflow ordering, and tool-call success. We use this score as a compact leaderboard metric, while still reporting the individual diagnostic components in the main results. The largest weights are assigned to tool selection and argument correctness because using the wrong scientific tool, spatial/temporal window, variable, threshold, or scenario parameter typically invalidates downstream Earth-science computations. Category coverage, workflow ordering, instruction validity, and execution success provide complementary diagnostics for whether the overall workflow is executable and scientifically coherent. To evaluate whether the heuristic weights materially affect our conclusions, we conduct two complementary analyses using the underlying item-level metric logs used to compute Table~\ref{tab:main_results}. 

\paragraph{Study A: predictive validity of process metrics.}
We first test whether the individual \textsc{ToolUseScore} components predict downstream answer success. For each model-item observation, we define a strict binary success label:
\begin{equation}
    \mathrm{Success} = \mathbf{1}\{\mathrm{Hit@tol} \ge 1\}.
\end{equation}
We fit a logistic regression with the process metrics as input features and \textsc{Success} as the dependent variable. Because \textsc{ToolCallSuccessRate} is identical to \textsc{InstAcc} in the current metric files, we drop it from this regression to avoid perfect collinearity. We also run a linear regression using continuous \textsc{NumScore} as the dependent variable as a graded numeric-correctness check. These regressions are used as diagnostic construct-validity analyses, not as causal models or as a procedure for tuning the official weights.

The logistic regression achieves an AUC of $0.803$ and McFadden pseudo-$R^2$ of $0.160$, showing that the process metrics used in \textsc{ToolUseScore} carry meaningful predictive signal for downstream answer success.

\begin{table}[t]
\centering
\small
\renewcommand{\arraystretch}{1.08}
\caption{\textbf{Predictive validity of process metrics.}
We predict item-level answer success from the process components used in \textsc{ToolUseScore}. \textsc{ToolCallSuccessRate} is omitted from the logistic regression because it is collinear with \textsc{InstAcc} in the current metric reports.}
\label{tab:tooluse_predictive_validity}
\begin{tabular}{llc}
\toprule
\textbf{Analysis} & \textbf{Quantity} & \textbf{Value} \\
\midrule
Logistic regression on $\mathbf{1}\{\mathrm{Hit@tol}\ge1\}$ & AUC & 0.803 \\
Logistic regression on $\mathbf{1}\{\mathrm{Hit@tol}\ge1\}$ & McFadden pseudo-$R^2$ & 0.160 \\
\bottomrule
\end{tabular}
\end{table}

\paragraph{Study B: sensitivity to alternative weighting schemes.}
We next test whether model rankings depend strongly on the exact heuristic weights. Starting from the published weights, we recompute model-level process scores under several alternative weighting schemes: equal weights across all components; leave-one-component-out weights, where each component is dropped once and the remaining weights are renormalized; regression-derived weights from Study A; and 1,000 random perturbations of the published weights by $\pm 0.10$, clipped to nonnegative values and renormalized. For each alternative scheme, we rebuild the model leaderboard and compute Spearman rank correlation against the original published-weight leaderboard.

The rankings are highly stable. Table~\ref{tab:tooluse_weight_sensitivity} show across 1,000 random perturbations, the median Spearman correlation with the published leaderboard is $0.996$, the minimum correlation is $0.982$, and every perturbation has $\rho \ge 0.95$. The top-ranked and bottom-ranked models remain unchanged in all perturbations. This indicates that the main process-level conclusions are not an artifact of one particular weighting choice.

\begin{table}[t]
\centering
\small
\renewcommand{\arraystretch}{1.08}
\caption{\textbf{Sensitivity of model rankings to \textsc{ToolUseScore} weights.}
We perturb the published weights by $\pm 0.10$, clip to nonnegative values, renormalize, and recompute the model leaderboard over 1,000 trials. Rank correlations are computed against the published-weight leaderboard.}
\label{tab:tooluse_weight_sensitivity}
\begin{tabular}{lc}
\toprule
\textbf{Quantity} & \textbf{Value} \\
\midrule
Random perturbation trials & 1,000 \\
Median Spearman $\rho$ & 0.996 \\
Minimum Spearman $\rho$ & 0.982 \\
Fraction with $\rho \ge 0.95$ & 1.000 \\
Top-ranked model unchanged & 1.000 \\
Bottom-ranked model unchanged & 1.000 \\
\bottomrule
\end{tabular}
\end{table}

These analyses support \textsc{ToolUseScore} as a stable model-level process summary. They do not imply that the published weights are uniquely optimal, nor do they eliminate the need to inspect individual diagnostic components. Rather, they show that the main process-level model rankings are robust across reasonable weighting schemes and that the process components are predictive of strict downstream answer success.

\subsection{Answer Evaluation and Unified NumScore}
\label{app:answer_metrics}

The model must return a machine-readable \texttt{<final\_json>} prediction. The answer sidecar is first validated against a strict schema requiring each field to contain exactly \texttt{key}, \texttt{value}, \texttt{abs\_tol}, \texttt{rel\_tol}, and \texttt{floor\_scale}. For numeric fields, let $y$ be the ground-truth value, $\hat{y}$ the prediction, and define the effective tolerance as
\begin{equation}
    \lambda = \max(\tau_{\mathrm{abs}}, \tau_{\mathrm{rel}} |y|, f),
\end{equation}
where $\tau_{\mathrm{abs}}$ is the absolute tolerance, $\tau_{\mathrm{rel}}$ is the relative tolerance, and $f$ is a field-specific floor scale. We compute
\begin{equation}
    AE = |\hat{y} - y|, \qquad n = \frac{AE}{\lambda}.
\end{equation}

\paragraph{Hit@tol.}
\begin{equation}
    \mathrm{Hit@tol}_{\mathrm{field}} =
    \begin{cases}
        1, & \text{if } AE \le \lambda, \\
        0, & \text{otherwise.}
    \end{cases}
\end{equation}

\paragraph{NumScore.}
\begin{equation}
    \mathrm{NumScore}_{\mathrm{field}} =
    \begin{cases}
        1, & \text{if } n \le 1, \\
        2^{-(n-1)}, & \text{if } n > 1.
    \end{cases}
\end{equation}
For multi-field outputs, both Hit@tol and NumScore are averaged over aligned ground-truth fields. Fields are aligned by key whenever keys are unique and match on both sides; otherwise alignment falls back to field order. Missing or unparseable numeric predictions receive zero credit. For non-numeric ground-truth fields, canonical exact matching is used, and the resulting direct-match value is used for both Hit@tol and NumScore.

\section{Bootstrap Confidence Intervals and Pairwise Comparisons}
\label{app:bootstrap_results}

We estimate finite-benchmark uncertainty using nonparametric bootstrap resampling over benchmark items. Tool steps, trace entries, and output fields are nested within an item and are not treated as independent samples. For each model and metric, we sample $N$ items with replacement, recompute the mean score, and repeat this procedure for $B=2000$ bootstrap replicates. We report percentile-based 95\% confidence intervals. For model comparisons, we bootstrap paired item-level differences. Intervals that exclude zero are treated as statistically reliable at the benchmark level.

\begin{table}[t]
\centering
\footnotesize
\renewcommand{\arraystretch}{1.08}
\caption{\textbf{Bootstrap confidence intervals for full-benchmark aggregate metrics.} Reported values are item-level bootstrap means and 95\% confidence intervals over the full benchmark ($N=403$ items).}
\label{app:tab_bootstrap_full_benchmark}
\resizebox{\linewidth}{!}{%
\begin{tabular}{llccc}
\toprule
\textbf{Model} & \textbf{Metric} & \textbf{N} & \textbf{Mean} & \textbf{95\% CI} \\
\midrule
Claude\_sonnet4.6 & ToolUseScore & 403 & 0.5922 & [0.5746, 0.6091] \\
Claude\_sonnet4.6 & NumScore     & 403 & 0.2844 & [0.2503, 0.3208] \\
Claude\_sonnet4.6 & Hit@tol      & 403 & 0.2288 & [0.1923, 0.2645] \\
\midrule
claude-haiku-4-5-20251001 & ToolUseScore & 403 & 0.4814 & [0.4625, 0.4994] \\
claude-haiku-4-5-20251001 & NumScore     & 403 & 0.2119 & [0.1783, 0.2450] \\
claude-haiku-4-5-20251001 & Hit@tol      & 403 & 0.1523 & [0.1224, 0.1847] \\
\midrule
gemini-25-flash & ToolUseScore & 403 & 0.3620 & [0.3422, 0.3817] \\
gemini-25-flash & NumScore     & 403 & 0.0770 & [0.0556, 0.1006] \\
gemini-25-flash & Hit@tol      & 403 & 0.0634 & [0.0426, 0.0872] \\
\midrule
gpt-5.5 & ToolUseScore & 403 & 0.4501 & [0.4294, 0.4711] \\
gpt-5.5 & NumScore     & 403 & 0.2645 & [0.2296, 0.3002] \\
gpt-5.5 & Hit@tol      & 403 & 0.2103 & [0.1734, 0.2445] \\
\midrule
qwen3.5:35b & ToolUseScore & 403 & 0.3995 & [0.3811, 0.4178] \\
qwen3.5:35b & NumScore     & 403 & 0.0749 & [0.0545, 0.0976] \\
qwen3.5:35b & Hit@tol      & 403 & 0.0589 & [0.0409, 0.0789] \\
\midrule
qwen3.5:9b & ToolUseScore & 403 & 0.3118 & [0.2929, 0.3316] \\
qwen3.5:9b & NumScore     & 403 & 0.0134 & [0.0048, 0.0249] \\
qwen3.5:9b & Hit@tol      & 403 & 0.0126 & [0.0048, 0.0234] \\
\bottomrule
\end{tabular}%
}
\end{table}

\begin{table}[t]
\centering
\footnotesize
\renewcommand{\arraystretch}{1.08}
\caption{\textbf{Pairwise bootstrap differences for a key model comparison.} Reported values are item-level paired bootstrap mean differences and 95\% confidence intervals over the full benchmark ($N=403$ items).}
\label{app:tab_bootstrap_pairwise}
\begin{tabular}{llccc}
\toprule
\textbf{Comparison} & \textbf{Metric} & \textbf{N} & \textbf{Mean Diff} & \textbf{95\% CI} \\
\midrule
Claude\_sonnet4.6 $-$ qwen3.5:35b & ToolUseScore & 403 & 0.1927 & [0.1718, 0.2143] \\
Claude\_sonnet4.6 $-$ qwen3.5:35b & NumScore     & 403 & 0.2095 & [0.1716, 0.2499] \\
Claude\_sonnet4.6 $-$ qwen3.5:35b & Hit@tol      & 403 & 0.1698 & [0.1320, 0.2097] \\
\bottomrule
\end{tabular}
\end{table}

\begin{figure}[t]
    \centering
    \begin{minipage}[t]{0.46\textwidth}
        \centering
        \includegraphics[width=\linewidth]{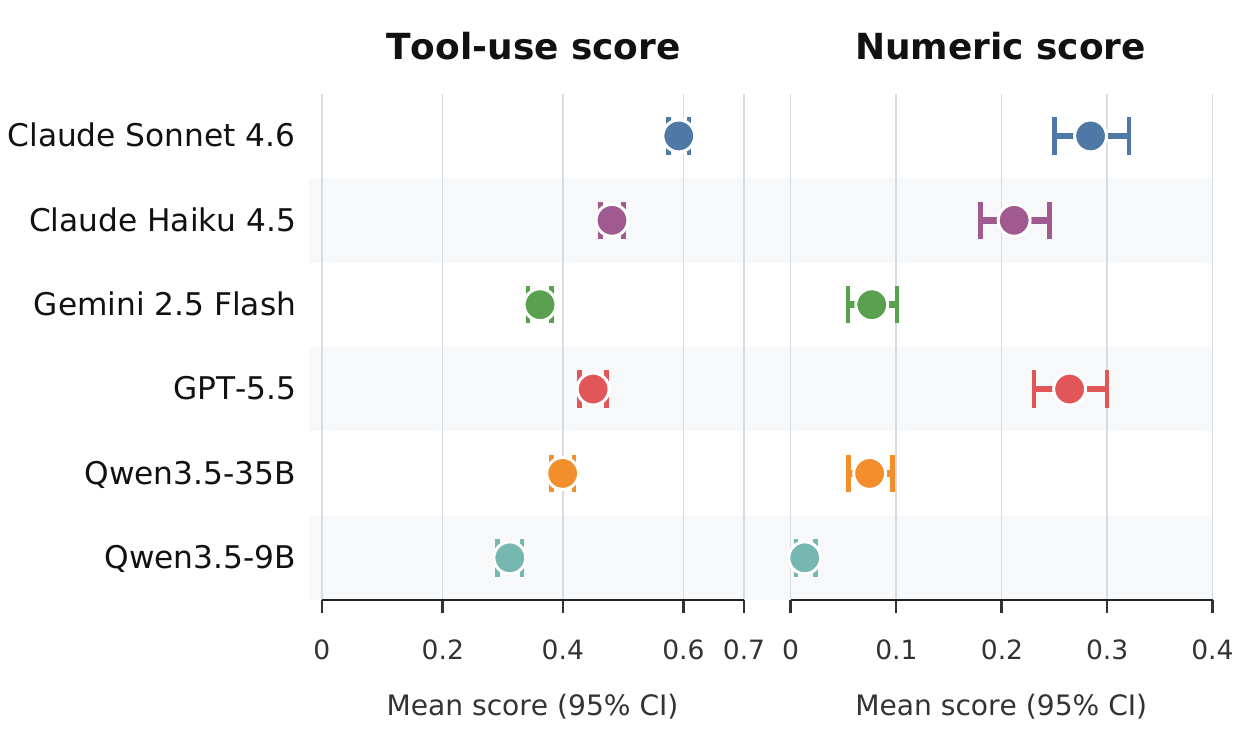}
        {\small (a)}
    \end{minipage}
    \hfill
    \begin{minipage}[t]{0.50\textwidth}
        \centering
        \includegraphics[width=\linewidth]{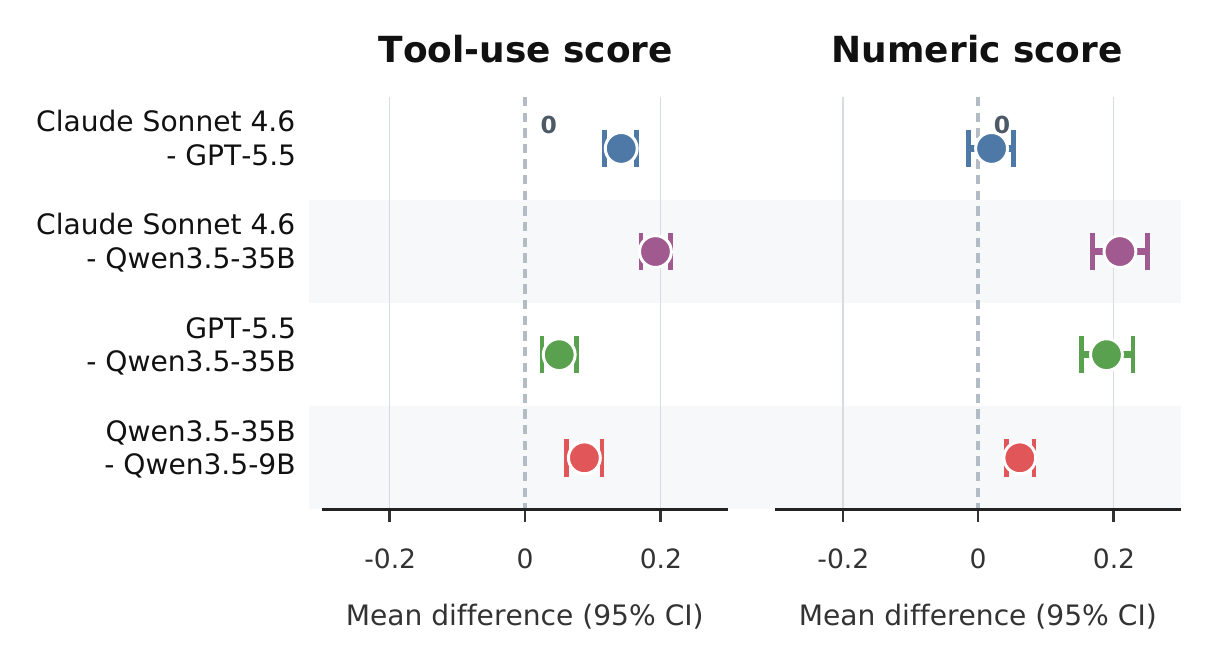}
        {\small (b)}
    \end{minipage}
    \caption{\textbf{Bootstrap confidence intervals for aggregate benchmark performance.} (a) Mean ToolUseScore and NumScore with 95\% item-level bootstrap confidence intervals on the full benchmark. (b) Paired mean differences with 95\% bootstrap confidence intervals for key model comparisons. Positive values indicate that the first-listed model outperforms the second.}
    \label{app:fig_bootstrap_ci}
\end{figure}

These intervals support full-benchmark aggregate comparisons at the observed effect sizes, especially for ToolUseScore. For example, Claude Sonnet 4.6 exceeds Qwen3.5-35B by 0.1927 ToolUseScore with a 95\% CI of [0.1718, 0.2143]. At the same time, smaller answer-level deltas and narrow subgroup analyses should be interpreted cautiously.

\section{Extended Failure-Mode Analysis}
\label{app:error_analysis}

\begin{figure}[t]
    \centering
    \includegraphics[width=0.92\linewidth]{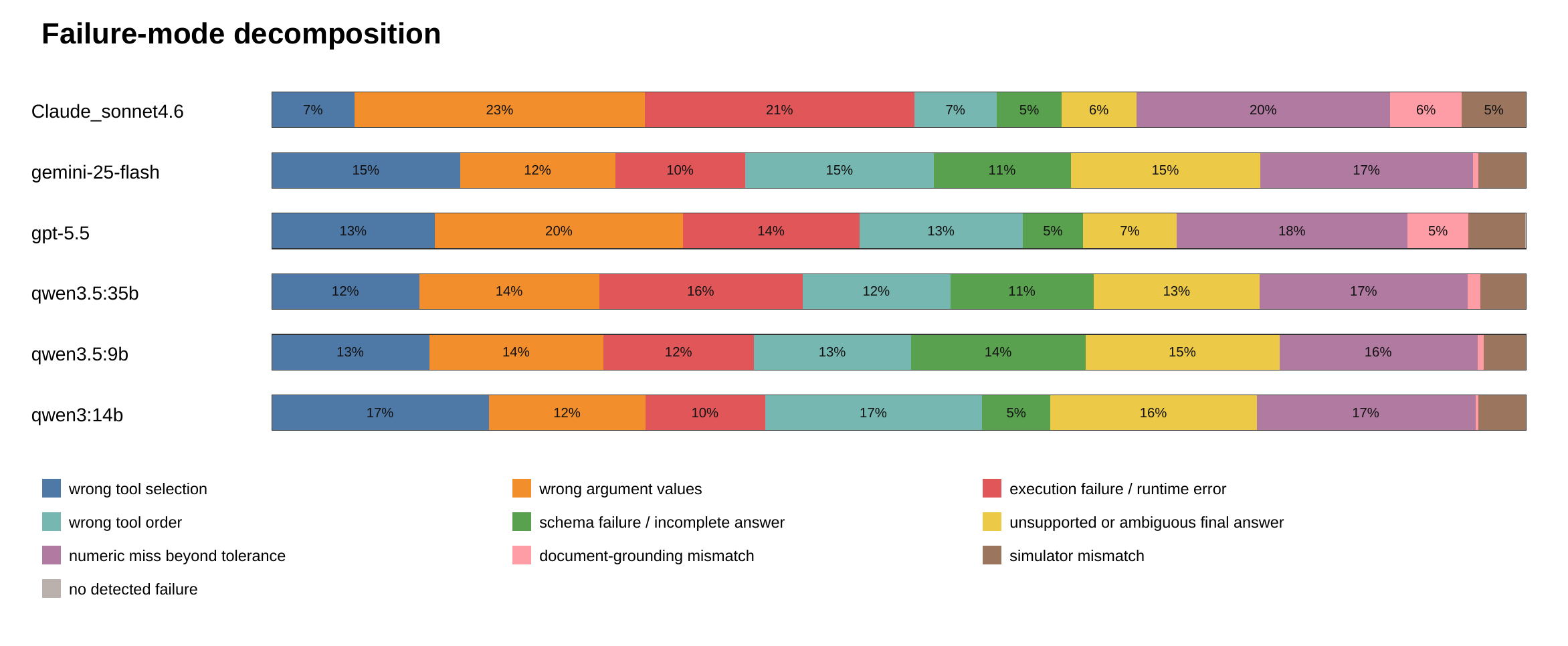}
    \caption{\textbf{Extended failure-mode decomposition.} Share of detected failure categories across representative models. Categories include tool-selection failures, argument errors, execution/runtime errors, ordering mistakes, schema or answer-format failures, numeric misses beyond tolerance, document-grounding mismatch, simulator mismatch, and no detected failure.}
    \label{app:fig_failure_modes}
\end{figure}

Figure~\ref{app:fig_failure_modes} decomposes failures into tool-selection, parameterization, execution, ordering, schema, answer-format, numeric-grounding, document-grounding, and simulator categories. Argument errors and numeric misses are the most pervasive failure modes, indicating that even when models select roughly appropriate tools, they often fail at specifying correct parameters, aligning retrieval windows or thresholds, and producing quantitatively accurate outputs. Tool-orchestration failures remain widespread, especially for weaker models, showing that reliable Earth-science agents need more than tool access: they need coherent workflow planning, precise argument grounding, and final answer construction that remains faithful to retrieved evidence and generated artifacts.

\section{Inference Execution and Computational Overhead}
\label{app:inference_time}

\begin{table}[t]
\centering
\caption{Inference execution statistics across evaluated frontier and open-weight models on the benchmark. Times represent wall-clock duration, including API latency, tool execution, and token generation.}
\label{app:tab:inference_time}
\resizebox{0.8\textwidth}{!}{
\begin{tabular}{lrrr}
\toprule
\textbf{Model} & \textbf{Runs} & \textbf{Total Inference} & \textbf{Avg/Run} \\
\midrule
claude-haiku-4-5-20251001 & 502 & 3d 15:47:34 & 00:10:30 \\
Claude\_sonnet4.6         & 442 & 5d 15:29:50 & 00:18:24 \\
gemini-2.5-flash          & 416 & 2d 03:38:00 & 00:07:27 \\
gemini-3.1-pro-preview    & 420 & 5d 14:50:58 & 00:19:16 \\
gemma4:26b                & 467 & 8d 15:51:37 & 00:26:42 \\
gemma4:e4b                & 335 & 18:12:20    & 00:03:16 \\
gpt-5.4                   & 410 & 2d 17:57:22 & 00:09:39 \\
gpt-5.5                   & 445 & 4d 16:22:28 & 00:15:09 \\
InternVL3                 & 431 & 03:38:31    & 00:00:30 \\
llama3.1:8b               & 469 & 01:44:59    & 00:00:13 \\
Mistral-7B-Instruct-v0.3  & 426 & 22:30:14    & 00:03:10 \\
qwen3-vl:8b               & 422 & 5d 13:03:33 & 00:18:55 \\
qwen3.5:35b               & 426 & 5d 09:37:42 & 00:18:15 \\
qwen3.5:9b                & 433 & 3d 02:29:06 & 00:10:19 \\
qwen3:1.7b                & 404 & 03:12:59    & 00:00:29 \\
qwen3:14b                 & 406 & 1d 06:20:07 & 00:04:29 \\
qwen3:4b                  & 406 & 2d 03:49:57 & 00:07:40 \\
qwen3:8b                  & 413 & 16:05:58    & 00:02:20 \\
\bottomrule
\end{tabular}
}
\end{table}

To contextualize the computational demands of evaluating agents on complex Earth-science workflows, Table \ref{app:tab:inference_time} details the execution statistics for both frontier and open-weight models. Because TerraBench tasks are not static text completions, these runtimes encompass end-to-end wall-clock execution, including model generation, dynamic tool invocation, geographic data retrieval, simulator overhead (e.g., AquaCrop, CLIMADA), and intermediate artifact parsing.

The data reveals a stark bifurcation in execution behavior. Frontier-class models and large reasoning-focused models exhibit substantial evaluation times, reflecting their tendency to engage in deep, multi-step ReAct loops, extended error correction, and comprehensive tool usage. Claude Sonnet 4.6 and Gemini 3.1 Pro Preview similarly averaged 18 to 19 minutes per run, indicating persistent, long-horizon workflow orchestration.

Conversely, compact open-weight models (e.g., Llama 3.1 8B, InternVL3, and Qwen3:1.7B) complete tasks in fractions of a minute. However, as demonstrated in our main quantitative results (Table~\ref{tab:main_results}), this speed is largely symptomatic of premature workflow termination, failure to utilize the available tool registry, and an inability to recover from intermediate execution errors. Ultimately, these execution metrics further highlight the unique difficulty of TerraBench: achieving high final-answer correctness requires models to sustain coherent, computationally expensive execution traces over highly extended time horizons.

\section{Computational Resources}\label{app:computational_resources}

Building and evaluating TerraBench required substantial computational resources due to the multi-step nature of the agentic workflows and the use of frontier models for trace generation and verification. The annotation pipeline consumed a total of 2,469,318,457 tokens, while the inference phase across all evaluated models consumed 820,028,249 tokens. API-based models were evaluated through their respective provider endpoints. All experiments with open-source models were conducted on a local node with 4 NVIDIA RTX A6000 GPUs.

We report inference time in Table~\ref{app:tab:inference_time} as the sum of per-question run durations, computed from each trace's recorded \texttt{started\_at} and \texttt{finished\_at} timestamps. This excludes idle wall-clock gaps introduced when model batches were manually paused and later resumed. Across the current experiment outputs, the timed inference runs sum to approximately 62.8 days of active agent execution across 8,216 runs; one empty trace file was excluded from this total.

\section{Experimental Setup}\label{app:experimental_setup}

We evaluate models in a standardized batch-inference setting. Each TerraBench item is represented as a row containing a context, question, and structured output specification. During evaluation, the batch runner constructs one agent request per row using only these fields, executes the TerraAgent workflow, and stores the resulting trace in a question-indexed directory (\texttt{Q1}, \texttt{Q2}, \ldots). Each run produces a complete trajectory file, a dialog trace, generated artifacts, and an extracted \texttt{<final\_json>} sidecar used for numerical evaluation.

All models are evaluated with high reasoning effort, same agent scaffold, tool interface, prompt template, and output contract. Tools are executed in local mode unless otherwise specified, and tool artifacts are written to the corresponding question directory to preserve reproducibility. For interrupted runs, we resume from the latest trace in the question directory rather than restarting the full batch.

Evaluation is performed by pairing prediction traces with ground-truth traces at the question-folder level. We report both task-answer quality and process-level behavior, including instruction accuracy, exact and relaxed tool accuracy, argument accuracy, tool-category F1, ordering score, tool-call success rate, file-boundary accuracy, guardrail compliance, structured answer validity, numerical tolerance accuracy, and text-similarity metrics. Numeric answers are evaluated from the extracted \texttt{number\_prediction.json} files against the corresponding ground-truth JSON fields using each item's absolute and relative tolerances.

To ensure a fair and reproducible comparison across model scales, frontier proprietary models are accessed via their respective official API endpoints. Conversely, all open-source models are hosted locally using Ollama~\cite{Lin_Safi_2025_ollama} on a dedicated compute node equipped with 4 NVIDIA RTX A6000 GPUs. Both API-based and local models are explicitly configured for high reasoning effort, utilizing maximum output token limits and appropriate sampling parameters to prevent premature generation truncation during complex, long-horizon Earth-science workflows.

\section{Appendix: Simulator Descriptions}
\label{app:simulator}

\subsection{Purpose}
TerraAgent includes a suite of simulator tools designed to translate climate, weather, hazard, exposure, and infrastructure inputs into sectoral impact estimates. These tools enable agentic workflows to autonomously select relevant model families, prepare compatible input files, execute reproducible impact analyses, and compare baseline and counterfactual scenarios under matched assumptions.

The simulator tools span five primary impact domains:
\begin{itemize}
    \item crop-water and agricultural production;
    \item physical asset and natural-hazard risk;
    \item building energy demand and thermal discomfort;
    \item road-network disruption and mobility impacts;
    \item human thermal stress and exposure-response health burden.
\end{itemize}

Additionally, the suite includes utility tools for preparing weather, exposure, and hazard-overlay artifacts required by the downstream simulators. All simulator executions return a structured response containing summary metrics, derived statistics, artifact file paths, provenance metadata, and execution warnings.

\subsection{Tool Inventory}

\begin{table}[hbt!]
\centering
\renewcommand{\arraystretch}{1.2}
\caption{Inventory of TerraAgent Simulator Tools.}
\resizebox{\textwidth}{!}{%
\begin{tabular}{l l p{6cm} p{4.5cm}}
\toprule
\textbf{Tool} & \textbf{Domain} & \textbf{Main role} & \textbf{Main outputs} \\
\midrule
\texttt{impact\_aquacrop\_run} & Crop-water & Estimates crop yield, biomass, ET, and water stress. & Yield, biomass, seasonal ET, daily crop-water trace. \\
\texttt{impact\_dssat\_run} & Crop systems & Estimates crop yield, water-stress days, and phenology. & Yield, water-stress days, seasonal ET, archived run. \\
\texttt{impact\_climada\_run} & Hazard-risk & Estimates event losses using intensity, exposure, and impact. & Expected impact, worst-event impact, risk curve. \\
\texttt{impact\_energyplus\_run} & Building & Estimates energy demand and discomfort under weather. & Total energy, peak demand, discomfort hours. \\
\texttt{impact\_sumo\_run} & Mobility & Estimates traffic disruption metrics. & Avg/p95 travel time, total delay, cancellations. \\
\texttt{impact\_health\_utci} & Thermal-stress & Computes UTCI-based stress metrics. & Mean UTCI, p95 UTCI, category counts, time series. \\
\texttt{impact\_health\_erf} & Health burden & Applies an exposure-response function. & Attributable fraction, baseline/attributable cases. \\
\bottomrule
\end{tabular}%
}

\label{tab:simulator_tools}
\end{table}

\subsection{Standardized Execution and Provenance}
All simulator tools follow a unified execution schema. Each run requires a \texttt{run\_id}, \texttt{scenario\_name}, scenario description, explicit assumptions, initialization seeds, resource metadata, and a timeout threshold. Outputs are written to isolated, tool-specific run directories. The tool's response explicitly records input artifacts, output artifacts, SHA-256 file hashes, content types, and runtime logs.

This schema ensures scientific reproducibility by making input and output paths explicit and hashable, allowing any result to be traced back to the exact files utilized by the agent. For counterfactual analysis, TerraAgent is expected to execute the same tool consecutively using matched settings: one for the baseline scenario and one for the intervention scenario. This isolates impact metrics to the specific variables altered between runs, ensuring rigorous scenario comparison.

\subsection{Domain-Specific Method Descriptions}

\subsubsection{Crop-Water and Agricultural Production}
\texttt{impact\_aquacrop\_run} executes an AquaCrop-style crop-water workflow. It ingests daily weather forcing, soil parameters, crop parameters, and management settings to calculate seasonal yield, biomass accumulation, evapotranspiration, irrigation demands, and water stress indices through daily water-balance computations.

\texttt{impact\_dssat\_run} represents a DSSAT-style crop systems workflow. It accepts dynamic weather, soil profiles, cultivar specifications, and management/treatment inputs to estimate crop yield driven by growing-degree-day accumulation, water stress, treatment effects, and phenology. The module outputs detailed day-level diagnostics alongside archived run folders for comprehensive analysis.

\subsubsection{Hazard Risk and Exposure}
\texttt{extract\_exposure\_layers} and \texttt{hazard\_overlay\_osm} process exposure and disruption files from local OpenStreetMap (OSM)-like extracts, hazard layers, population data, and satellite-derived summaries. These data-preparation tools map local asset valuation, road exposure, and hazard-to-network intersections required by downstream risk and mobility models.

\texttt{impact\_climada\_run} executes a CLIMADA-style physical risk workflow. It intersects hazard event sets with exposure records using specific vulnerability or impact functions. The model calculates deterministic damage estimates, returning asset-level losses, total event losses, expected annual impacts, and risk exceedance curves.

\subsubsection{Building Energy and Thermal Discomfort}
\texttt{weather\_to\_epw} constructs standard EPW-style hourly weather forcing files from daily or hourly meteorological tables. \texttt{impact\_energyplus\_run} then applies these weather files alongside building design specifications (IDF format) to evaluate building energy performance. The simulator calculates hourly total, cooling, and heating loads based on floor area, dry-bulb temperature, operational setpoints, and fixed sensitivity parameters, returning comprehensive demand and thermal discomfort metrics.

\subsubsection{Road-Network Disruption}
\texttt{impact\_sumo\_run} processes SUMO-style traffic disruption workflows. It ingests transportation networks, trip route definitions, and disruption parameters (e.g., edge closures, speed reductions). By evaluating edge-level travel-time penalties and re-routing or canceling trips across disrupted infrastructure, the simulator reports average travel times, 95th-percentile delays, overall network throughput, and trip cancellation rates.

\subsubsection{Human Health}
\texttt{impact\_health\_utci} evaluates human thermal-stress exposure from meteorological time series. By processing temperature, humidity, wind speed, and mean radiant temperature, it computes the Universal Thermal Climate Index (UTCI) to categorize physiological stress, identifying periods of strong to extreme heat or cold stress.

\texttt{impact\_health\_erf} estimates the attributable health burden resulting from environmental exposures using established exposure-response functions (ERFs). It integrates exposure time series, exposed population counts, baseline incidence rates, and specific beta coefficients (with optional confidence intervals and lag structures) to compute relative risk, attributable fractions, and total attributable health cases.

\subsection{External Documentation Sources}
These sources define the external model families and scientific methodologies that the TerraAgent simulator tools emulate:
\begin{itemize}
    \item FAO AquaCrop: \url{https://www.fao.org/aquacrop/en/}
    \item DSSAT tools and data overview: \url{https://dssat.net/tools-data-overview/}
    \item CLIMADA documentation: \url{https://climada-python.readthedocs.io/}
    \item EnergyPlus documentation: \url{https://energyplus.readthedocs.io/}
    \item Eclipse SUMO documentation: \url{https://eclipse.dev/sumo/docs/}
    \item UTCI homepage: \url{https://utci.org/}
    \item EPA BenMAP-CE: \url{https://www.epa.gov/benmap}
    \item OpenStreetMap copyright and license: \url{https://www.openstreetmap.org/copyright/en}
\end{itemize}

\section{Partial Annotator Agreement on Trace Acceptance}
\label{app:annotator_agreement}

To assess the reproducibility of our trace-acceptance protocol, we conducted an inter-annotator agreement audit on a random subset of 50 benchmark candidates from the annotation pool. Two independent annotators reviewed each item under the strict trace-acceptance criteria used during benchmark construction. These criteria included workflow fidelity, provenance support, document/execution separation, schema compliance, tolerance validity, leakage control, answer completeness, and trace honesty. Crucially, the unit of agreement was the item-level accept/reject decision rather than isolated agreement over individual trace steps or output fields.

\begin{table}[H]
\centering
\small
\renewcommand{\arraystretch}{1.2}
\caption{\textbf{Partial Inter-annotator agreement audit.} Agreement is computed on item-level accept/reject decisions under the canonical trace-acceptance rubric.}
\begin{tabular}{lc}
\toprule
\textbf{Quantity} & \textbf{Value} \\
\midrule
Items audited & 50 \\
Sampling & Random subset of benchmark \\
Raw agreement & 88.0\% \\
Cohen's $\kappa$ & 0.694 \\
Disagreements & 6 \\
\bottomrule
\end{tabular}
\label{tab:annotator_agreement}
\end{table}

We measured inter-rater reliability using Cohen's $\kappa$ score~\cite{cohen}. The audit yielded an 88.0\% raw agreement rate and a Cohen's $\kappa$ of $0.694$. In standard statistical interpretations, the score indicates substantial agreement, which is highly encouraging given the multi-step complexity of the tasks and the strictness of the rubric. While this audit does not replace a full independent double-annotation over all 403 finalized items, it provides strong preliminary evidence that the benchmark's trace-acceptance protocol is reproducible and that the gating decisions are reliable.